\documentclass{article}
\usepackage{float}
\usepackage[margin=1.0in]{geometry}
\usepackage{abstract}

\usepackage[dvipsnames,svgnames, table,xcdraw]{xcolor}
\usepackage{relsize}
\usepackage[most]{tcolorbox}
\usepackage{listings}
\usepackage{subcaption}
\usepackage{algorithm}
\usepackage{algorithmic}
\usepackage{comment}
\usepackage{graphicx}
\usepackage{booktabs}
\usepackage{multirow}
\usepackage{enumitem}
\usepackage{url}
\usepackage{mathpazo}
\usepackage{amsmath,amssymb,amsthm,bm}
\usepackage[toc,page,header]{appendix}
\usepackage{fancyhdr}
\usepackage[backend=biber,style=nature,natbib=true,maxbibnames=99,minalphanames=3]{biblatex}
\usepackage[colorlinks=true]{hyperref}
\usepackage[T1]{fontenc}
\usepackage{cleveref}
\usepackage{adjustbox}
\usepackage{longtable}

\usepackage{auth_detailed}

\counterwithout{figure}{section}
\counterwithout{table}{section}
\definecolor{darkgoldenrod}{rgb}{0.72, 0.53, 0.04}
\definecolor{backgroundcolor}{RGB}{250, 250, 252}   
\definecolor{keywordcolor}{RGB}{30, 0, 178}       
\definecolor{stringcolor}{RGB}{204, 0, 102}        
\definecolor{numbercolor}{RGB}{0, 128, 128}        
\definecolor{emphcolor}{RGB}{30, 0, 178}            
\definecolor{commentcolor}{RGB}{0, 128, 0}       
\definecolor{basiccodecolor}{RGB}{61, 61, 61}

\lstdefinestyle{customstyle}{
    backgroundcolor=\color{backgroundcolor},   
    commentstyle=\color{commentcolor},
    keywordstyle=\color{keywordcolor},
    numberstyle=\color{numbercolor},
    stringstyle=\color{stringcolor},
    basicstyle=\color{basiccodecolor}\ttfamily\footnotesize,
    breakatwhitespace=false,         
    breaklines=true,                 
    captionpos=b,                    
    keepspaces=true,                 
    numbers=left,     
    basicstyle=\color{basiccodecolor}\ttfamily\footnotesize,
    numbersep=5pt,             
    xleftmargin=2em,
    xrightmargin=2em,
    showspaces=false,                
    showstringspaces=false,
    showtabs=false,                  
    tabsize=1,
    frame=single,
    framesep=5pt,
    framexleftmargin=1.5em,
    framexrightmargin=1.5em,
    framextopmargin=1pt,
    framexbottommargin=1pt,
    aboveskip=10pt,
    belowskip=10pt,
    breaklines=true,
    breakautoindent=true,
    emph={textgrad, tg, Variable, MultipleChoiceTestTime,
    TextualGradientDescent, BlackboxLLM},             %
    emphstyle={\color{emphcolor}},
    extendedchars=true,
}

\definecolor{logocolor}{RGB}{30, 0, 178}

\definecolor{darkerlogocolor}{RGB}{20, 0, 145}  

\newtcolorbox{ttcolorbox}[1][]{colframe=darkerlogocolor, colback=darkerlogocolor!4!white, title=#1}

\newtcolorbox{apxtcolorbox}[1][]{colframe=black, colback=black!3!white, title=#1}

\definecolor{ForestGreen}{RGB}{34,139,34} 
\definecolor{RoyalBlue}{RGB}{65,105,225}
\definecolor{TitleColor}{HTML}{B7B2D0}
\definecolor{ContentColor}{HTML}{DBD8E7}

\newcommand{\ie}{\em{i.e.}}
\newcommand{\eg}{\em{e.g.}}
\newcommand{\platform}[1]{ChatBattery}

\newcommand{\NMCInput}{LiNi\textsubscript{0.8}Mn\textsubscript{0.1}Co\textsubscript{0.1}O\textsubscript{2}}
\newcommand{\NMCSiMg}{LiNi\textsubscript{0.7}Mn\textsubscript{0.05}Co\textsubscript{0.05}Si\textsubscript{0.1}Mg\textsubscript{0.1}O\textsubscript{2}}
\newcommand{\NMCSiCa}{LiNi\textsubscript{0.65}Mn\textsubscript{0.1}Co\textsubscript{0.1}Si\textsubscript{0.1}Ca\textsubscript{0.05}O\textsubscript{2}}
\newcommand{\NMCMgB}{LiNi\textsubscript{0.65}Mn\textsubscript{0.1}Co\textsubscript{0.1}Mg\textsubscript{0.1}B\textsubscript{0.05}O\textsubscript{2}}

\hypersetup{
  colorlinks=true,
  urlcolor=BrickRed,
  citecolor=RoyalBlue,
  linkcolor=BrickRed
} 
\BiblatexSplitbibDefernumbersWarningOff
\title{Expert-Guided LLM Reasoning for Battery Discovery: From AI-Driven Hypothesis to Synthesis and Characterization}

\author{\name Shengchao Liu$^{1}$\thanks{Equal contribution.}, 
\name Hannan Xu$^{2}$\textsuperscript{*},
\name Yan Ai$^{1}$\textsuperscript{*},\quad
\name Huanxin Li$^{2,3}$\textsuperscript{ †},
\name Yoshua Bengio$^{1,4,5}$\textsuperscript{ †},
\name Hongyu Guo$^{6,7}$ \thanks{Corresponding authors. Email: hongyu.guo@uottawa.ca, yoshua.bengio@mila.quebec, huanxin.li@ucl.ac.uk}   \\ \\
$^{1}$Universit\'e de Montr\'eal, Canada \\
$^{2}$University of Oxford, UK\\
$^{3}$University College London, UK\\
$^{4}$Mila - Qu\'ebec AI Institute, Canada \\
$^{5}$CIFAR AI Chair, Canada \\
$^{6}$University of Ottawa, Canada\\
$^{7}$National Research Council Canada\\
}

\begin{document}

\fancyhead[R]{\platform{}}
\setlength{\headheight}{13pt}
\pagestyle{fancy}

\newcounter{suppfigure}
\newcounter{supptable}
\makeatletter
\newcommand\suppfigurename{Supplementary Figure}
\newcommand\supptablename{Supplementary Table}
\newcommand\suppfigureautorefname{\suppfigurename}
\newcommand\supptableautorefname{\supptablename}
\let\oldappendix\appendix
\renewcommand\appendix{%
    \oldappendix
    \setcounter{figure}{0}%
    \setcounter{table}{0}%
    \renewcommand\figurename{\suppfigurename}%
    \renewcommand\tablename{\supptablename}%
}
\makeatother

\maketitle

\renewcommand{\abstractnamefont}{\normalfont\normalsize\bfseries}
\renewcommand{\abstracttextfont}{\normalfont\normalsize}

\renewcommand{\abstractname}{Abstract}
\begin{abstract}
\abstracttextfont
Large language models (LLMs) leverage chain-of-thought (CoT) techniques to tackle complex problems, representing a transformative breakthrough in artificial intelligence (AI). However, their reasoning capabilities have primarily been demonstrated in solving math and coding problems, leaving their potential for domain-specific applications—such as battery discovery—largely unexplored. Inspired by the idea that reasoning mirrors a form of guided search, we introduce \platform{}, a novel agentic framework that integrates domain knowledge to steer  LLMs toward more effective reasoning in materials design. Using \platform{}, we successfully identify, synthesize, and characterize three novel lithium-ion battery cathode materials, which achieve practical capacity improvements of 28.8\%, 25.2\%, and 18.5\%, respectively, over the widely used cathode material, \NMCInput{} (NMC811). Beyond this discovery, \platform{} paves a new path by showing a successful LLM-driven and reasoning-based platform for battery materials invention. This complete AI-driven cycle—from design to synthesis to characterization—demonstrates the transformative potential of AI-driven reasoning in revolutionizing materials discovery.
\end{abstract}

\section{Introduction}
Artificial intelligence (AI) has recently sparked a new wave of problem-solving paradigms, driving transformative advancements across various scientific fields. This surge in AI applications is unlocking unprecedented productivity in scientific discovery, with notable breakthroughs in areas such as protein structure prediction~\citep{Jumper2021AlphaFold} and weather forecasting~\citep{Lam2024GenCast}. These innovations demonstrate AI's potential to not only deepen our understanding of complex systems but also accelerate the discovery process in areas once constrained by traditional methods, unlocking new frontiers in research and innovation. 
However, the application of AI in materials design—especially in harnessing the reasoning capabilities of large language models (LLMs) for developing battery materials to tackle critical open-ended challenges like climate change—has remained relatively underexplored. 

\textbf{AI for Battery Materials Discovery.} 
The accelerating global energy transition, which is driven by electric vehicles and large-scale energy storage, demands next-generation lithium-ion batteries with significantly higher energy density. Battery materials discovery is a complex and multidisciplinary endeavor that requires the integration of computational modeling, materials synthesis, and experimental characterization. Traditional methods often rely on time-consuming trial-and-error approaches, limiting the pace of innovation. Recent advancements in AI have empowered approaches for knowledge-driven hypothesis generation for materials. For instance, AI has been utilized to screen over 32 million candidates to discover new materials with the potential for better batteries~\citep{Chen_2024}. AI has also demonstrated its capability for the autonomous optimization of non-aqueous Li-ion battery electrolytes, successfully identifying several fast-charging formulations~\citep{Dave_2022}. These AI-driven efforts for material discovery mainly fall into  two categories: (1) Utilize physics-inspired geometric models for energy prediction~\citep{liu2023symmetry}, which can be further combined for material molecule dynamics simulation~\citep{merchant2023scaling}; (2) Utilize large language models (LLMs) as agents for experimental automation~\citep{boiko2023autonomous}. However, the reasoning capability of LLMs for material design has not been thoroughly explored.

\begin{figure}[t]
    \centering
    \includegraphics[width=\linewidth]{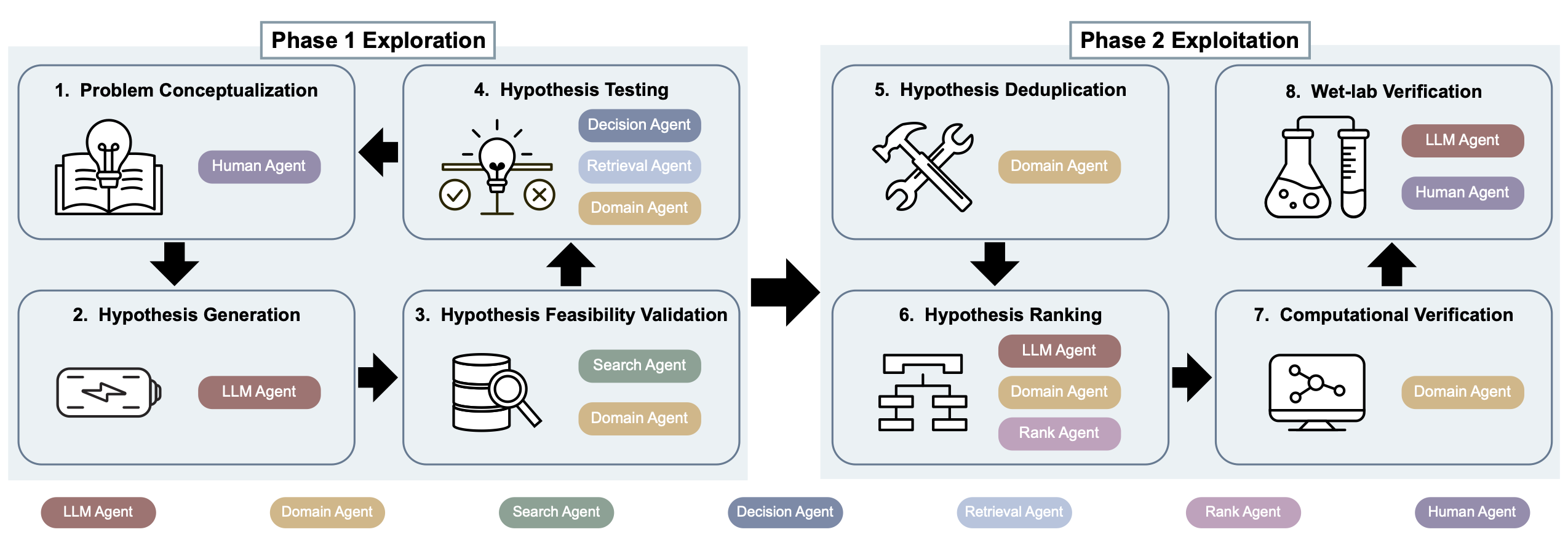}
    \vspace{-2ex}
    \caption{\small 
   \textbf{Overview  of the \platform{} pipeline.}
    It includes two main phases, i.e., Exploration (left) and Exploitation (right), eight sequential stages (rounded rectangles), and seven specialized agents (legend at  bottom). 
    }
    \label{fig:01_pipeline}
\end{figure}

\textbf{Large Reasoning Model for Scientific Discovery.} Reasoning in Large Language Models (LLMs), including the Large Reasoning Models (LRMs), has emerged as a transformative advancement in artificial intelligence~\citep{NEURIPS2020_1457c0d6,wei2022chain}. Inspired by human reasoning processes, these models facilitate the identification of novel ideas and solutions. Notable examples include OpenAI's o1/o3-mini~\citep{openai2025o3mini}, DeepSeek's R1~\citep{deepseekai2025deepseekr1incentivizingreasoningcapability}, Grok3 Reasoning, and Gemini Flash Thinking~\citep{gemini_flash_thinking}, all of which have demonstrated exceptional capabilities in logical inference and problem-solving. Integrating reasoning capabilities is crucial for accelerating scientific discoveries, as it enables efficient search mechanisms at the heart of hypothesis generation and testing, reducing reliance on traditional trial-and-error methodologies~\citep{liu2023ChatDrug,gottweis2025towards,liu2023MoleculeSTM,liu2025ProteinDT}. However, in domain-specific applications like battery materials discovery, current models often fall short, producing incomplete or unreliable reasoning traces. Additionally, while methods such as supervised fine-tuning (SFT)~\citep{NEURIPS2020_1457c0d6}  and low-rank adaptation (LoRA)~\citep{hu2021lora} can help adapt LRMs for more effective reasoning, they typically rely on access to reasoning trace datasets—which can be limited in availability and costly to curate in specialized domains like battery materials. Reinforcement learning-based approaches for enhancing LLM reasoning also tend to demand substantial computational resources during training. 

To address the aforementioned challenges, in this paper, we introduce \platform{}, an expert-guided approach to leverage LLM reasoning capabilities for battery materials discovery. \platform{} aims to correctly and sufficiently reason out promising hypotheses for the discovery of new materials. By incorporating expert knowledge directly into the reasoning process, we ensure that the hypotheses generated by large language models (LLMs) are both scientifically valid and relevant, reducing the risk of errors and enhancing the effectiveness of the discovery process. This expert-guided reasoning provides a critical layer of oversight, ensuring that the AI's reasoning behaviors and traces are effective and grounded in domain knowledge. To showcase the benefits of \platform{}, we have leveraged \platform{} to effectively  identify, synthesize, and characterize three novel lithium-ion battery cathode materials~\footnote{Cathodes play a critical role in lithium-ion batteries, largely determining key performance metrics such as voltage and energy density.}: \NMCSiMg{} (NMC-SiMg), \NMCSiCa{} (NMC-SiCa), and \NMCMgB{} (NMC-MgB), derived from the widely used lithium battery cathode \NMCInput{} (NMC811). Compared to NMC811, which has a specific capacity of about 135 mAh/g, the three candidates show reversible capacities of 174, 169, and 160 mAh/g by the third cycle. These represent improvements of 28.8\%, 25.2\%, and 18.5\%, respectively.

\textbf{Our Contributions.}
In this work, we introduce \platform{}, a novel expert-guided LLM reasoning platform for battery materials discovery, as illustrated in~\Cref{fig:01_pipeline}. Notably, (1) \platform{} adopts the reasoning capabilities of large language models (LLMs) by integrating reasoning guidance from domain experts into multiple agents to effectively tackle complex battery materials design challenges. (2) Using \platform{}, we effectively identified, synthesized, and characterized three novel lithium-ion battery cathode materials, achieving meaningful real-world capacity improvements over the widely used NMC811 cathode.\looseness=-1

To the best of our knowledge, \platform{} represents the first successful real-world implementation of LLM-driven battery materials design encompassing the full design-to-synthesis-to-characterization cycle. \platform{}'s AI-driven approach enables a rapid discovery and optimization process, significantly reducing the time required for experimental screening and validation.

\begin{figure}[t]
    \centering
    \includegraphics[width=\linewidth]{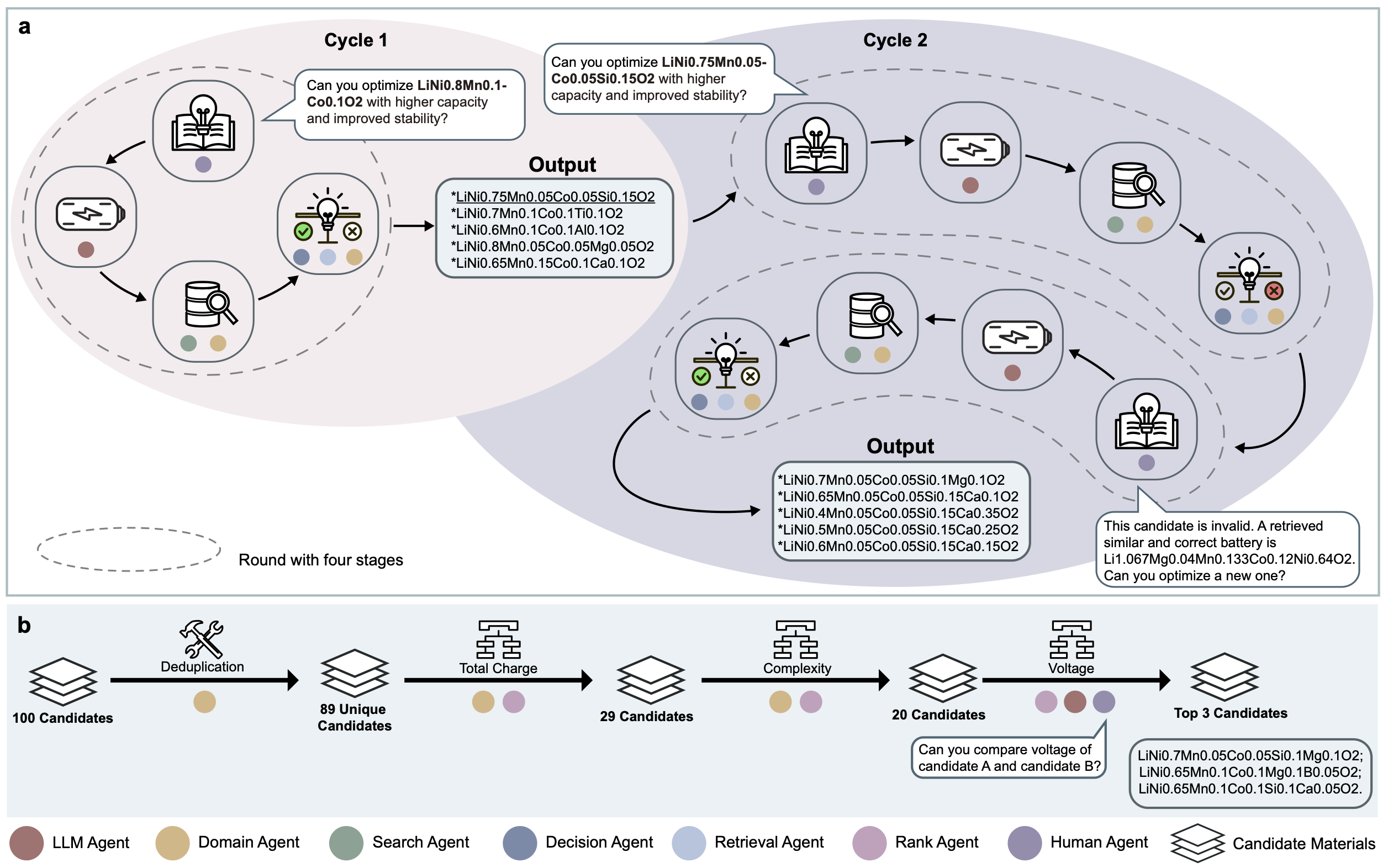}
    \vspace{-2ex}
    \caption{\small 
\textbf{Instantiation of \platform{}.} (a) Illustrates two cycles of Phase 1 in \platform{}, from stage 1 to stage 4; Cycle 2 builds upon Cycle 1’s output and demonstrates the use of the Retrieval Agent. (b) Depicts stages 5 and 6 of Phase 2, where top candidate materials are progressively identified, evaluated, and selected  (from left to right). 
    }
    \label{fig:02}
\end{figure}

\section{\platform{} Platform}
\platform{} is an AI-driven material optimization platform structured into two synergistic phases: exploration and exploitation. Together, these phases encompass eight sequential stages, orchestrated by seven specialized agents. The full workflow is illustrated in \Cref{fig:01_pipeline} and detailed next. 

\subsection{Exploration and Exploitation}
\textbf{Phase 1 Exploration.} This phase focuses on extensively exploring potential optimization candidates by searching across the chemical space. It includes the following four stages. 
\begin{itemize}[noitemsep,topsep=0pt]
    \item \textbf{Stage 1 Problem Conceptualization.} The process begins with identifying and framing a scientific question or challenge to be addressed.
    \item \textbf{Stage 2 Hypothesis Generation.} At this stage, a potential solution or explanation is proposed. This hypothesis serves as a foundational framework for subsequent exploration.
    \item \textbf{Stage 3 Hypothesis Feasibility Evaluation.} This stage aims to evaluate whether the hypothesis, or a closely related one, has been previously reported in the literature and to assess its plausibility and viability for continued stages.
    \item \textbf{Stage 4 Hypothesis Testing.} This stage involves applying computational methods as surrogate functions to evaluate whether the hypothesis is supported by simulated evidence.
\end{itemize}
These four stages are repeated recursively until, for example, 100 candidates are generated, enabling a systematic exploration of a wide range of possibilities.

\noindent \textbf{Phase 2 Exploitation.} Phase 1 involves exploring a broad range of potential options, whereas the subsequent exploitation phase aims to refine the search by eliminating unlikely candidates and concentrating on the most promising ones.  It encompasses the following four stages. 
\begin{itemize}[noitemsep,topsep=0pt]
    \item \textbf{Stage 5 Hypothesis Deduplication.} After the exploration phase, we obtain a list of candidate materials. This stage filters out redundant materials by a domain agent.
    \item \textbf{Stage 6: Hypothesis Ranking.} All remaining candidates are ranked based on a composite score that incorporates three key factors: total charge, structural complexity, and predicted voltage. At each filtering stage, only the top candidates are selected to proceed to the next stage.
    \item \textbf{Stage 7: Computational Validation.} 
    DFT calculations are conducted to evaluate the stability and 
    properties of the top-ranked cathode candidates in silico.
    \item \textbf{Stage 8: Wet-lab Validation.} In the final stage, selected candidates are synthesized and tested through wet-lab experiments to validate their real-world performance.
\end{itemize}

\subsection{Supportive Agents} 
As shown in~\Cref{fig:01_pipeline}, seven agents are communicating with each other to support the key stages in the \platform{} platform. We provide a brief description in the main article below. Please check the Methods and Materials Section \cite{methods} for more details.
\begin{itemize}[noitemsep,topsep=0pt]
    \item \textbf{LLM agent} plays two key roles in \platform{}. (1) At stage 2, it generates optimized cathode material candidates by modifying input formulas according to specified constraints ({\eg}, capacity > 300 mAh/g and allowed elements). (2) At stage 6, it assists in qualitatively ranking the cathode candidates based on factors such as their thermodynamic stability.
    \item \textbf{Search agent} filters out cathode compounds  that already exist by querying domain-specific databases.
    \item \textbf{Decision agent} evaluates whether an optimized cathode candidate is valid by checking if its theoretical capacity exceeds that of the input compound. If not, it invokes the retrieval agent to provide domain-specific feedback.
    \item \textbf{Retrieval agent} retrieves a similar but valid compound from a database to guide further iterations when a generated cathode material is invalid.
    \item \textbf{Rank agent} evaluates cathode candidates using a hierarchical ranking tree that incorporates multiple factors. It prioritizes key criteria such as total charge, preparation complexity, and voltage to identify the most promising candidates for further validation.
    \item \textbf{Domain agent} provides domain-specific computations and functions to support other agents, including (1) Enabling exact and range-based matching for the search agent. (2) Computing capacity using a formula based on Li count and molecular weight for the decision and retrieval agent. (3) Measuring the similarity between chemical formulas based on element counts and predefined weights for the retrieval agent. (4) Ranking the generated batteries based on various pre-defined factors.
    \item \textbf{Human agent} provides expert judgment and oversight at critical stages of the \platform{}, including the prompt design in problem conceptualization and wet-lab verification.
\end{itemize}

\section{Cathode Material Discovery with \platform{}}
To highlight the significance of our \platform{} framework, we focus on optimizing a high-nickel lithium nickel manganese cobalt oxide (NMC) cathode material, NMC811.

\textbf{NMC811 (\NMCInput{})} is one of the most promising next-generation cathode materials and has been successfully commercialized, particularly in electric vehicles and large-scale energy storage systems~\citep{saaid2024ni,son2020systematic}. It offers high energy density and reduced cobalt dependency. However, challenges remain, including structural instability, limited cycling stability, and suboptimal capacity. To address these issues, we aim to use NMC811 optimization as the target task and test it with the \platform{} platform.

We provide, in \Cref{sec:phase_1} and \Cref{sec:phase_2}, an overview of the eight stages involved in NMC811 optimization. In addition, we offer detailed and extended discussions on the final two stages, namely computational and wet-lab verifications, in \Cref{sec:phase_2_computational_verification} and \Cref{sec:phase_2_wet_lab_verification}, respectively.

\subsection{\platform{} for NMC811 Optimization, Phase 1 -- Exploration} \label{sec:phase_1}

The first phase of \platform{} for NMC811 optimization is exploration. In this phase, we input the NMC811 material, using text prompts, into the \platform{} platform, which then conducts an extensive search for a wide range of potential optimization outcomes.

\textbf{Cycles and Rounds.} The entire optimization process is structured into cycles, with each cycle comprising multiple rounds, and each round spanning Stages 1 to Stage 4. (1) Within a single cycle, rounds are executed iteratively until a list of $k$ \textbf{valid} candidate batteries is obtained. This set of rounds constitutes one complete cycle. (2) After each cycle, the resulting $k$ candidates can serve as inputs for the next cycle. Thus, if each cycle generates $k$ candidates and the process runs for $C$ cycles, the total number of candidates produced is $k^C$. This entire exploration process can be repeated $N$ times, resulting in a total of $N \times k^C$ candidates. In our experiment, we set $k = 5$, $C = 2$, and $N = 4$, yielding 100 candidate cathode materials. These 100 candidates are then passed to the exploitation phase, as described in~\Cref{sec:phase_2}. We illustrate this process in Figure~\ref{fig:02} and detail it next. 

\textbf{Stage 1 Problem Conceptualization.} At this stage, human experts, acting as the Human Agent, design prompts for the LLM to optimize the properties of NMC811. The objective is to \textbf{enhance theoretical capacity} while maintaining material stability by incorporating new elements and adjusting the ratios of existing ones in NMC811. To account for synthesis cost and safety, the LLM is instructed to prioritize elements from the carbon group, alkaline earth metals, and transition metals, while avoiding radioactive elements. There are roughly three cases for prompt design: the prompt at the initial round of the initial cycle, the prompt at the initial round of the latter cycles, and the prompts at the latter rounds. Please check~\ref{sec:methods_and_materials}, namely Methods and Materials, for the detailed prompt design.

\textbf{Stage 2 Hypothesis Generation.} The resulting prompts from Stage 1 above are then fed into an LLM agent by \platform{} to perform the optimization task. In our main experiment, we use GPT-3.5 as the LLM agent. Given that we set $k = 5$, the LLM agent generates five novel cathode material candidates, which are then passed on to the next stage.

\textbf{Stage 3: Hypothesis Feasibility Validation.} This stage evaluates whether the optimized cathode materials have been previously reported. To do so, we leverage two databases (the Inorganic Crystal Structure Database (ICSD)~\citep{zagorac2019recent} and the Materials Project~\citep{jain2013commentary}) and employ a Search Agent for automated verification. Details of this process are provided in the Methods and Materials section \ref{sec:methods_and_materials}. If a generated material is found in either database, it will be flagged in Stage 4 and considered during prompt design (see \ref{sec:methods_and_materials}) in Stage 1 of the subsequent optimization round.

\textbf{Stage 4: Hypothesis Testing.} This stage assesses the quality of the optimized materials by evaluating their theoretical capacities, as determined by a Domain Agent. If the optimized compound exhibits a lower capacity than NMC811, the Decision Agent flags it as invalid and forwards it to the Retrieval Agent, which searches for a similar compound with higher capacity. This domain feedback is then incorporated into the prompt design (see \ref{sec:methods_and_materials}) for Stage 1 in the next round. If all $k$ optimized samples are valid, the current round is considered complete; otherwise, the process returns to Stage 1 to start the next round of optimization.

The resulting candidates from this exploration phase are then passed into the exploitation phase of the \platform{} pipeline, which will be discussed in detail next. 

\subsection{\platform{} for NMC811 Optimization, Phase 2 -- Exploitation} \label{sec:phase_2}
The second phase in \platform{} for NMC811 optimization is exploitation. After exploring a wide range of potential valid candidates in the first phase, this phase focuses on selecting the most promising candidates with the highest confidence. It concludes with a final wet-lab experiment, where the selected candidates are synthesized and characterized. In detail, this phase comprises the following four stages. 

\textbf{Stage 5 Hypothesis Deduplication.} After Phase 1 of the NMC811 optimization pipeline, we have 100 candidate materials. In this stage, we apply a range match to remove duplicate compounds from the generated list. As a result, 89 unique materials remain after the deduplication process.

\textbf{Stage 6: Hypothesis Ranking.} In this stage, the 89 candidate cathode materials are ranked according to three properties: total charge, preparation complexity, and total voltage. The most promising optimized materials, based on the ranking results, are then fed into the subsequent steps.

The first metric is the \textit{total charge}. It refers to the calculation of the net charge of the 89 generated compounds. The valences of Li, Mn, Co, Ni, and O are set to match those in NMC811 to ensure consistency and comparability in theoretical capacity and voltage. For other elements, their highest common oxidation states are used, excluding elements that exist primarily in various acid salt forms (\Cref{tab:valence_values_for_total_charge}). This is not a rigorous formal charge calculation but rather an approximation based on oxidation-state heuristics. The initial calculation identifies 11 compounds with a total charge of zero (\Cref{tab:SI_03}). However, since transition metals often exhibit variable oxidation states and valence deviations can occur during experimental synthesis, we broaden the selection criteria. We rank all 89 candidates by the absolute value of their total charge and select the 29 with the smallest values, {\ie}, those with values between -0.1e and 0.1e.

The second metric is the \textit{preparation complexity}. It refers to the number of different elements present in a compound, as not every reactant can be a single substance in the synthesis process. In general, a larger number of element types indicates a more complex synthesis route. Based on this metric, 9 candidates with more than seven elements are excluded, and the 20 with the lowest preparation complexity are selected.

The last metric is the \textit{voltage}, also known as the electric potential difference. It is the measure of the electric potential energy per unit charge between two points in an electric circuit. The precise calculation of voltage requires molecular dynamics (MD) simulation. Because of the immense computational cost of MD simulation, we use an LLM agent as a proxy to assess the voltage of the 20 filtered cathode materials from the last step. More concretely, instead of exact voltage calculation, an LLM is employed to perform qualitative voltage pairwise comparisons of materials, leveraging its reasoning abilities (prompts and other details are in the Methods and Materials Section, namely~\ref{sec:methods_and_materials}). The results are then processed using a merge sort to determine the ranking. We note that, upon expert review of all comparison documents and their associated reasoning traces, we  find the modeling's reasoning to be reliable,   though not exhaustive; notably, the LLM assessment reflects the influence of each element on the compound's voltage, particularly highlighting the role of nickel—higher nickel content generally correlates with a higher voltage~\citep{oliveira2024high}. Based on the final ranking, the top 3 out of 20 optimized cathode materials are selected for the subsequent stages: \NMCSiMg{}, \NMCMgB{}, \NMCSiCa{}. We denote them as NMC-SiMg, NMC-MgB, and NMC-SiCa, respectively. 

\textbf{Stage 7 Computational Verification.} At this stage, we employed density functional theory (DFT) calculations to assess the validity of the three selected materials. More details are discussed in~\Cref{sec:phase_2_computational_verification}.

\textbf{Stage 8 Wet-lab Verification.} The most compelling evidence for the effectiveness of using \platform{} in NMC811 optimization is the successful synthesis of the optimized materials. We have synthesized the optimized candidates, with further details provided in \Cref{sec:phase_2_wet_lab_verification}.

\begin{table}[t]
\centering
\caption{\small \textbf{DFT total energy.} For each supercell size, up to 1000 structures were sampled. } \label{tab:dft_energy}
\vspace{-2ex}
\resizebox{\textwidth}{!}{%
\begin{tabular}{l c l c}
\toprule
Candidate Formula & Supercell Size& Supercell Formula & Avg Total Energy (eV)\\
\toprule
LiNi\textsubscript{0.7}Mn\textsubscript{0.05}Co\textsubscript{0.05}Si\textsubscript{0.1}Mg\textsubscript{0.1}O\textsubscript{2}
     &2x2x2 &Li24Mg3Mn1Co1Si2Ni17O48&-502.5511$\pm$0.3916 \\     
     &2x2x1 &Li12Mg1Mn1Co1Si1Ni8O24& -255.9692$\pm$0.2775 \\
     & & & \\   LiNi\textsubscript{0.65}Mn\textsubscript{0.1}Co\textsubscript{0.1}Mg\textsubscript{0.1}B\textsubscript{0.05}O\textsubscript{2}
    &2x2x2 &Li24Mg2Mn3Co2Ni16B1O48&-506.3595$\pm$0.4430\\
     &2x2x1 &Li13Mg1Mn1Co1Ni7B1O24& -251.0930$\pm$0.4422\\
    & & & \\        LiNi\textsubscript{0.65}Mn\textsubscript{0.1}Co\textsubscript{0.1}Si\textsubscript{0.1}Ca\textsubscript{0.05}O\textsubscript{2}
     &2x2x2 &Li24Ca1Mn2Co2Si3Ni16O48&-515.4147$\pm$0.2862\\
     &2x2x1 &Li13Ca1Mn1Co1Si1Ni7O24&-252.4021$\pm$0.4643 \\
\bottomrule
\end{tabular}
}
\end{table}

\subsection{Computational Verification} \label{sec:phase_2_computational_verification}

\textbf{Initial Positions.}
The three initial CIF documents of NMC-SiMg, NMC-MgB, and NMC-SiCa are prepared based on NMC811 (Collection Code: 60827), downloaded from the ICSD dataset~\citep{zagorac2019recent}. The three CIF documents contain the following elements: Li, Mn, Co, Ni, Si, Mg, O; Li, Mn, Co, Ni, Mg, B, O; and Li, Mn, Co, Ni, Si, Ca, O, respectively. All three materials share the same space group: R-3m (Space Group No. 166). 

The introduced elements, Si, Mg, B, and Ca, partially substitute the original elements Mn, Co, Ni, or Li at the (0, 0, 0) site. This setting is because NMC811 has a layered structure in which the transition metal cations, Ni, Mn, and Co, occupy specific sites within the crystal lattice. The (0, 0, 0) site corresponds to the metal cation site in the lattice, as specified in the original NMC811 CIF document. The introduced elements have ability to maintain or slightly modify the structural integrity of the NMC811 lattice, and are chemically compatible with the existing transition metal cations in the structure. The occupancies of these introduced elements are adjusted accordingly to achieve full occupation of the site (occupancy sum = 1), since maintaining a fully occupied site is crucial for preserving the structural integrity of the lattice and preventing defects that could compromise material stability. Furthermore, using the formula ratios of these cations to set site occupancies ensures that the crystal structure accurately reflects the composition of the material.

The full CIF files are presented in the supplementary materials and have also been uploaded to the ICSD database.

\textbf{Partial Occupancy.} 
As observed from their formulas, all three candidates identified above are materials with partial occupancy, placing them in the category of disordered crystals. Site mixing in disordered materials presents two key challenges for property computation. First, standard methods such as density functional theory (DFT) require fully specified atomic positions, making it difficult to apply them directly. Second, unlike ordered crystals, disordered structures permit multiple occupancy possibilities at the same lattice sites, leading to an exponentially large number of potential configurations. 

\textbf{Supercells for DFT Calculation.} 
To address the aforementioned challenges, we adopt a widely used strategy among domain experts: expanding partially occupied candidate structures into larger supercells with fully occupied configurations and then sampling possible atomic configurations. For this purpose, we use the Supercell package~\citep{supercell}. This supercell expansion transforms partially occupied structures into fully occupied configurations, making them suitable for DFT calculations. Specifically, we generate two supercell configurations: 2×2×2 and 2×2×1, sampling up to 1000 structures for each size. This supercell expansion, however, results in a new issue. That is, the large atomic size of these supercell configurations, combined with the need for extensive sampling, makes direct evaluation using conventional methods such as ab initio molecular dynamics (AIMD) or density functional theory (DFT) with VASP computationally prohibitive. To address this problem, we employ machine learning surrogate models to predict the energies and forces of the sampled supercell structures. 

While several foundation models have been developed for small molecules and materials~\citep{choudhary2021atomistic,Mishin_2021,Zhang_2018,Zuo_2020,Chen_2022,deng_2023_chgnet}, we employ MACE-MP~\citep{batatia2023foundation} to evaluate the supercell structures derived from the top three candidates identified by the \platform{} pipeline. MACE-MP was selected due to its high accuracy in modeling interatomic interactions in periodic systems, its ability to generalize across different material classes, and its competitive performance on both force and energy prediction benchmarks. Its efficiency and reliability make it especially suitable for evaluating large, disordered supercells where traditional DFT becomes computationally prohibitive.

\textbf{Computational Results.} 
For each candidate material, we compute the total energy of each of the sampled supercells using MACE-MP and report the averaged total energy for each supercell size. The results are summarized in~\Cref{tab:dft_energy}.

The results in \Cref{tab:dft_energy} show that all three candidate formulas exhibit negative total energies, indicating that they are energetically favorable overall. This suggests that these compositions are thermodynamically stable enough to exist in principle, providing a strong starting point for further experimental synthesis and characterization, as will be discussed next in~\Cref{sec:phase_2_wet_lab_verification}. We note that while negative total energy alone does not guarantee synthesizability or long-term phase stability, it is a necessary condition that supports the feasibility of these candidate materials. 

\begin{figure}[t]
    \centering
    \includegraphics[width=.9\linewidth]{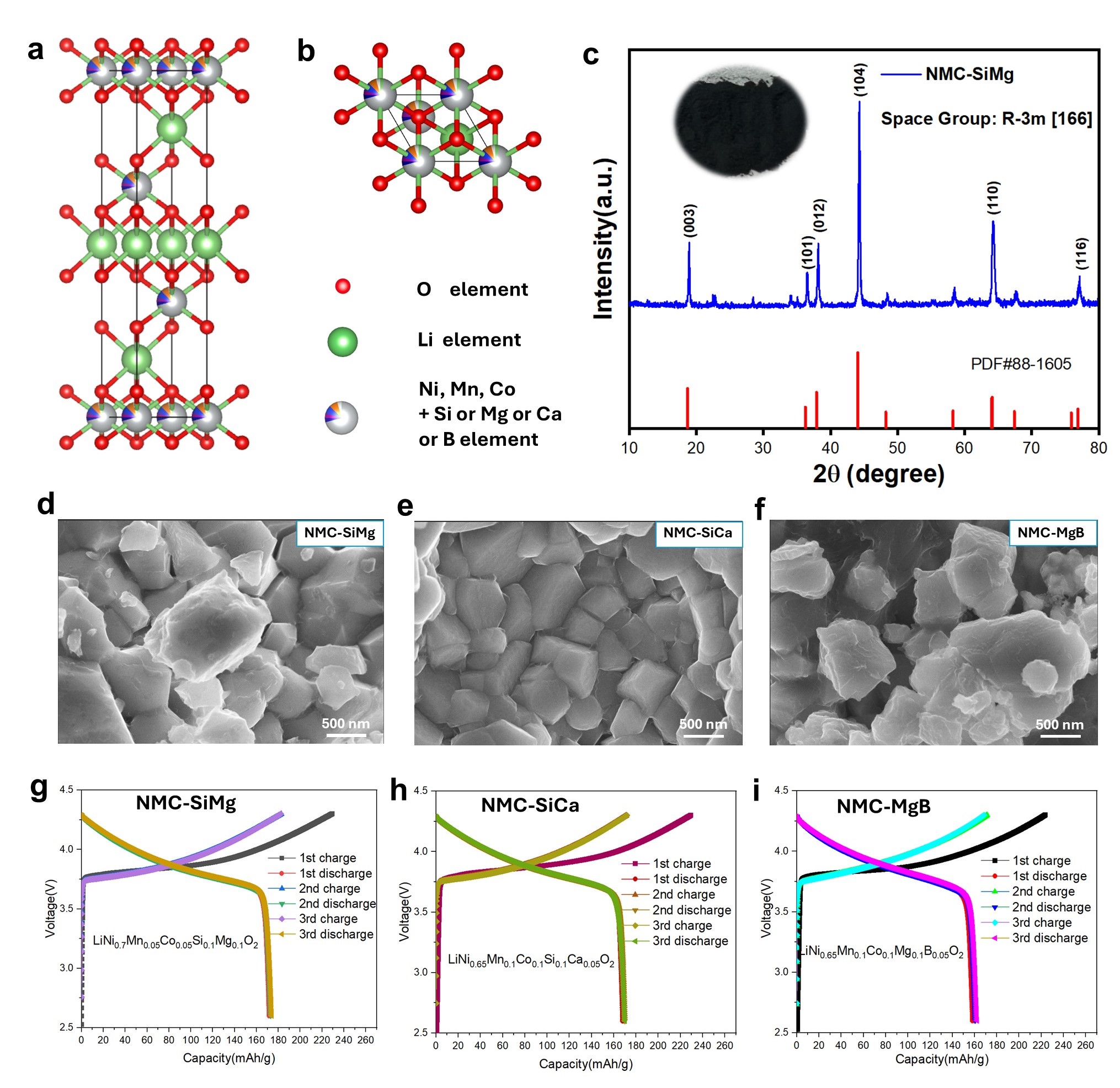}
    \vspace{-3ex}
    \caption{\small \textbf{Visualization of Structural and Electrochemical Characterization for NMC-Si/Mg/Ca/B Composites.} (a) The side view and (b) the top view of crystal structure for NMC-SiMg, NMC-SiCa, and NMC-MgB; (c) XRD pattern for NMC-SiMg; (d) SEM image of NMC-SiMg; (e) SEM image of NMC-SiCa; (f) SEM image of NMC-MgB; The first 3 cycle charge discharge profiles for (g) NMC-SiMg, (h) NMC-SiCa, and (i) NMC-MgB.
    }
    \label{fig:wet-lab-validation}
\end{figure}

\subsection{Wet-Lab Experimental Synthesis and Characterization} \label{sec:phase_2_wet_lab_verification}

The three candidate cathode materials predicted by \platform{}, namely NMC-SiMg, NMC-SiCa, and NMC-MgB, were successfully synthesized. The first material prepared was NMC-SiMg, with the synthesis procedure of all the NMC811-derived samples detailed in~\Cref{sec:wet_lab_synthesis}.

The crystal structure of NMC-SiMg, shown as a representative example in~\Cref{fig:wet-lab-validation}a, exhibits a typical layered architecture analogous to that of conventional NMC cathodes, with partial substitution of the transition metal sites by Si and Mg dopants. The top-view projection of the structure (\Cref{fig:wet-lab-validation}b) confirms the retention of a rhombohedral symmetry with the same space group (R-3m, No. 166) as NMC811. The XRD pattern in~\Cref{fig:wet-lab-validation}c further supports this structural similarity, matching the PDF No. 88-1605. However, a notable difference is observed in the preferred orientation, with the dominant diffraction peak shifting from the (003) plane, commonly seen in NMC811, to the (104) plane, indicative of a more disordered structure~\citep{pender2020electrode}. The inset in~\Cref{fig:wet-lab-validation}c presents an optical image of NMC-SiMg, revealing a characteristic fine black powder typical of LIB cathode materials.

Additionally, minor unidentified impurity phases are detected, as evidenced by weak diffraction peaks near 2$\theta$ = 22°, 28°, and 30°, which are absent in standard NMC811. The presence of these impurities is reasonable, considering this represents the first successful synthesis of the NMC-SiMg cathode guided by LLMs predictions. Such impurities may have been introduced at various stages of the synthesis process and can likely be minimized through further optimization of the fabrication protocol.

Following the successful synthesis of NMC-SiMg, the other two materials were also synthesized using the same procedure, as detailed in~\Cref{sec:wet_lab_synthesis}. SEM images of NMC-SiMg, NMC-SiCa, and NMC-MgB (\Cref{fig:wet-lab-validation}d–f) reveal that all three materials exhibit a rhombohedral particle morphology with sizes ranging from 500 to 800 nm. Notably, NMC-SiMg and NMC-MgB display slightly rougher surface textures, whereas NMC-SiCa shows a smoother surface, suggesting a comparatively higher degree of crystallinity.

The electrochemical performance of these materials is shown in the charge–discharge profiles in~\Cref{fig:wet-lab-validation}g–i, measured within a voltage window of 2.6–4.3 V. All three candidates exhibit stable cycling behavior with an average discharge voltage of approximately 3.85 V, significantly higher than that of conventional LIB cathode materials (typically 3.4–3.7 V)~\citep{mizushima1980lixcoo2,yuan2011development}. This enhanced voltage is consistent with the reasoning outputs generated by~\platform{}. Relative to the baseline NMC811, which delivered a specific capacity of around 135 mAh/g under identical testing conditions, the predicted candidates, NMC-SiMg, NMC-SiCa, and NMC-MgB, demonstrated superior reversible capacities of 174, 169, and 160 mAh/g, respectively, by the third cycle. These represent improvements of 28.8\%, 25.2\%, and 18.5\%, respectively, highlighting the potential of LLM-guided design in advancing cathode material performance (\Cref{tab:wetlab}).

\begin{table}[t]
\centering
\caption{\small \textbf{The comparison of NMC811 and the three novel cathode materials.}} \label{tab:wetlab}
\vspace{-2ex}
\begin{tabular}{lcccc}
\toprule
 & NMC811 & NMC-SiMg & NMC-SiCa & NMC-MgB \\
\toprule
Structure & Layered & Layered & Layered & Layered \\
Theoretical Capacity (mAh/g) & 275.50 & 294.66 & 287.29 & 293.08\\
Available Capacity (mAh/g) & 135 & 174 & 169 & 160 \\

\bottomrule
\end{tabular}
\end{table}

\section{Discussion and Outlook}

LLMs are poised to dramatically accelerate the discovery of novel battery materials by enabling rapid hypothesis generation and exploration of vast chemical design spaces. In this study, we introduced \platform{}, an expert-guided framework that integrates domain-specific knowledge into the reasoning process of LLMs. This approach bridges the gap between the general-purpose capabilities of LLMs and the nuanced requirements of materials science, enabling more accurate, relevant, and experimentally actionable predictions. 

By embedding expert reasoning into the generative pipeline, \platform{} ensures that proposed materials are not only chemically plausible but also aligned with physical and electrochemical constraints critical to battery performance. We validated the efficacy of this framework through the discovery, synthesis, and electrochemical evaluation of three novel lithium-ion cathode materials: NMC-SiMg, NMC-SiCa, and NMC-MgB. All three candidates significantly outperformed the state-of-the-art NMC811 in terms of practical specific capacity, with improvements of 28.8\%, 25.2\%, and 18.5\%, respectively. Furthermore, these materials demonstrate enhanced voltage profiles and superior cycling stability, highlighting the practical relevance of \platform{}'s predictions. 

A key feature of \platform{} is its ability to provide transparent and rational explanations for its generated choices, effectively mirroring expert reasoning. This interpretability not only fosters confidence in the proposed candidates but also facilitates iterative refinement in collaboration with human scientists. Notably, the entire discovery pipeline—from initial hypothesis generation to material synthesis and performance validation—was completed within a few months, a process that typically spans several years using conventional experimental workflows. This rapid turnaround underscores the transformative potential of LLM-augmented discovery pipelines in accelerating the pace of scientific innovation.

While the results demonstrate the utility of \platform{} in advancing battery materials research, the current system does exhibit limitations. Despite the generation of chemically novel candidates, most predicted materials remain structurally and compositionally close to known classes. This suggests that \platform{}, in its present form, is more adept at optimizing within known paradigms than at generating fundamentally new chemistries. As such, expert input remains essential to expand the system's exploration boundaries and push beyond conventional chemical spaces. Importantly, this interplay between AI-driven generation and human-guided refinement also creates unexpected opportunities, as demonstrated in the refinement of AI-suggested materials into even more advanced cathode compositions.  However, advances anticipated with future reasoning AIs are likely to provide greater exploration and creativity. A good example of this direction is the use of amortized inference methods such as GFlowNets that are focused on rich exploration of the space of possibilities~\citep{bengio2021flow,jain2022biological,jain2023gflownets,hu2024amortizing}.

\textbf{Beyond the AI-Predicted Candidates.} 
Among the three successful candidates generated by \platform{}, NMC-SiMg emerged as the most promising, demonstrating strong performance across multiple evaluation metrics. Building upon this foundation, domain experts further refined the material through targeted wet-lab experimentation, resulting in the development of a more advanced cathode: Li-rich-NMC-SiMg. Notably, this new material achieved a 34\% practical capacity improvement over NMC811, marking a significant step forward in lithium-ion battery performance. Although Li-rich-NMC-SiMg was not directly predicted by the AI model, its discovery underscores the powerful synergy between AI-guided design and expert-driven innovation. Full details of its synthesis and characterization are provided in~\citep{Characterization,Synthesis}.

Looking forward, the modular and domain-agnostic design of \platform{} renders it applicable well beyond battery research. Its core methodology—combining LLMs with expert-in-the-loop reasoning—can be readily adapted to other materials domains, including catalysts, semiconductors, and structural materials. Moreover, its integration into biology and environmental science holds the promise of accelerating discovery in diverse fields where complex, multivariate systems require both broad search capacity and deep domain insight. 

In conclusion, \platform{} represents a compelling advancement in AI-assisted scientific discovery. It exemplifies how human-AI collaboration can overcome current limitations of generative models and achieve accelerated, high-quality outcomes in materials research. As the framework matures, further developments aimed at enhancing its creative autonomy and domain adaptability will be crucial to unlocking its full potential as an AI co-scientist across disciplines.

\clearpage
\section{Methods and Materials} \label{sec:methods_and_materials}

\subsection{LLM Agent and Human Agent in Prompt Design}

\textbf{Stage 1 Problem Conceptualization.} In the main article, we mentioned that there are three cases for the prompt design, each corresponding to the process at different rounds and cycles. The modified phrases are highlighted in \textbf{bold}.

\begin{enumerate}[noitemsep,topsep=0pt]
\item \textbf{Initial round of initial cycle}. The prompt is:
\small
\begin{tcolorbox}[colback=ContentColor, colframe=TitleColor, title=Prompt for Initial Round of the Initial Cycle]
We have a Li cathode material \textbf{\{PLACEHOLDER\}}. Can you optimize it to develop new cathode materials with higher capacity and improved stability?\\
You can introduce new elements from the following groups: \textbf{carbon group, alkaline earth metals group, and transition elements}, excluding radioactive elements; and incorporate new elements directly into the chemical formula, rather than listing them separately; and give the ratio of each element; and adjust the ratio of existing elements.\\
My requirements are proposing five optimized battery formulations, listing them in bullet points (in asterisk *, not - or number or any other symbol), ensuring each formula is chemically valid and realistic for battery applications, and providing reasoning for each modification.
\end{tcolorbox}
\normalsize
Here, ``\textbf{\{PLACEHOLDER\}}'' is the placeholder for the input formula of the cathode materials.

\item \textbf{Subsequent rounds (initial \& subsequent cycle)}. In the first round (Stage 1 to 4), the LLM agent may generate invalid results, marked by our \platform{} platform. Therefore, we conduct multiple rounds of \platform{} optimization until all $k$ valid optimized cathode materials are successfully identified. Given that the theoretical capacity of a cathode material is negatively correlated with its molecular weight, the prompt not only adjusts the element ratios of previously optimized materials and introduces new candidates but also directs the replacement of newly added elements with alternatives of lower atomic mass. The prompt for Stage 1 in the subsequent round is as follows:
\small
\begin{tcolorbox}[colback=ContentColor, colframe=TitleColor, title=Prompt for Subsequent Rounds (Initial \& Subsequent Cycles)]
You generated some existing or invalid battery compositions that need to be replaced with valid ones (one for each).\\
\textbf{
These batteries have been discovered before:\\
* \{PLACEHOLDER\}\\
* \{PLACEHOLDER\}\\
...\\
These invalid batteries are:\\
* \{PLACEHOLDER\} (a retrieved similar and correct battery is \{PLACEHOLDER\})\\
* \{PLACEHOLDER\} (a retrieved similar and correct battery is \{PLACEHOLDER\})\\
...\\
}
When replacing the invalid or existing compositions, you can \textbf{replace the newly added elements with elements of lower atomic mass; and adjust the ratio of existing elements; and introduce new elements}. The new compositions must be stable and have a higher capacity. The final outputs should include newly generated valid compositions, skip the retrieved batteries, and be listed in bullet points (in asterisk *, not - or number or any other symbol).
\end{tcolorbox}
\normalsize

\item \textbf{Initial round of the subsequent cycles}. In the initial optimization cycle, one input material formula yields $k$ output formulas. These outputs can then be iteratively optimized in subsequent cycles, with the prompt modified accordingly. Notably, incorporating elements from the same group does not significantly improve material properties, as their similar valence electron configurations lead to chemically similar bonds. Therefore, in the second cycle, elements from the same group as those already present in the first-cycle input are excluded. For example, if the initial input compound includes an \textit{element from the carbon group}, the prompt will be modified as follows:
\small
\begin{tcolorbox}[colback=ContentColor, colframe=TitleColor, title=Prompt for Initial Round of the Subsequent Cycles]
We have a Li cathode material \textbf{\{PLACEHOLDER\}}. Can you optimize it to develop new cathode materials with higher capacity and improved stability?\\
You can introduce new elements from the following groups: \textbf{carbon group, alkaline earth metals group, and transition elements}, excluding radioactive elements; and incorporate new elements directly into the chemical formula, rather than listing them separately; and give the ratio of each element; and adjust the ratio of existing elements.\\
You can introduce new elements from the following groups: \textbf{alkaline earth metals group, and transition elements}, excluding radioactive elements; and incorporate new elements directly into the chemical formula, rather than listing them separately; and give the ratio of each element; and adjust the ratio of existing elements.\\
My requirements are proposing five optimized battery formulations, listing them in bullet points (in asterisk *, not - or number or any other symbol), ensuring each formula is chemically valid and realistic for battery applications, and providing reasoning for each modification.
\end{tcolorbox}
\normalsize
\end{enumerate}
These prompts are then fed into a large language model (LLM) for material optimization. Specifically, we use GPT-3.5 with a frequency penalty of 0.2 and a temperature of 1. The goal is to generate $k = 5$ optimized batteries, each representing a distinct \textbf{hypothesis} in our context.

\textbf{Stage 6 Hypothesis Ranking.}
We also consider using the LLM agent at Stage 6 for hypothesis ranking. The objective is to incorporate multiple factors into the ranking process, as detailed below. For factors such as voltage, which are difficult to estimate using surrogate functions, the LLM agent provides an alternative means of assigning ranking scores. The prompt for using LLM as the voltage surrogate for ranking is as follows:
\small
\begin{tcolorbox}[colback=ContentColor, colframe=TitleColor, title=Prompt Voltage Ranking]
Could you compare the two Li cathode materials, \textbf{{PLACEHOLDER}} and \textbf{{PLACEHOLDER}}, and identify which one has a higher voltage vs. Li+/Li (V)?\\
List the better one in the last line, marked by '*'.
\end{tcolorbox}
\normalsize
We use GPT-o4 to provide detailed reasoning information. It is important to note that hallucinations remain an inherent limitation of large language models at this stage. Therefore, additional verification is conducted in later stages: 
DFT calculation in stage 7 and wet lab synthesis in stage 8.

Notably, in our \platform{} platform, we also enable users to adjust and modify the prompt based on their domain knowledge. Please check the website in the GitHub repository for more details.

\subsection{Search Agent}
We incorporate the searching agent to filter out generated cathode materials  that already exist. For a candidate material $x_{\text{output}}$ and a retrieval database $D$. If the searching agent finds $x_{\text{output}}$ in $D$, then we will reject this candidate compound. Here, we consider two databases, as detailed below.

\textbf{Materials Project API.} The first searching agent queries the Materials Project (marked as $D_{\text{MP}}$) via its Application Programming Interface (API) calls to determine whether the generated cathode material $x_{\text{output}}$ already exists in the database~\citep{jain2013commentary}.

\textbf{ICSD API.} The second searching agent interacts with the Inorganic Crystal Structure Database (ICSD) (marked as $D_{\text{MP}}$) to check whether the generated compound $x_{\text{output}}$ is already present~\citep{zagorac2019recent}. In this case, we have manually downloaded the ICSD data, 10,096 lithium materials with chemical formulas, and use the domain agent to perform a \textit{range match} (will be discussed in detail later) instead of an exact match.

\subsection{Decision Agent}
The Decision Agent aims to provide a verdict on whether the output candidate generated by \platform{} has an improved theoretical capacity compared to the given input material. 

In the proposed \platform{}, the battery's theoretical capacity is used as the primary criterion for determining the validity of an optimization. Specifically, given an input cathode material $x_{\text{input}}$ and an output candidate $x_{\text{output}}$, the decision is defined as follows:
\begin{equation} \label{eq:satisfication}
\text{decide} (x_{\text{input}}, x_{\text{output}}) =
\begin{cases}
    \text{True}, & \text{if }  \text{ Capacity}_{x_{\text{output}}} > \text{ Capacity}_{x_{\text{input}}}\\
    \text{False}, & \text{otherwise.}
\end{cases}
\end{equation}
That is, True here indicates that the theoretical capacity of the output candidate generated by \platform{} is improved compared to that of the input material. 

The theoretical capacity function is provided by the domain agent, as will be introduced below.

\subsection{Retrieval Agent}
The Retrieval Agent aims to retrieve materials for the invalid  compounds generated. The retrieved materials are required to have a higher theoretical capacity compared to the input materials and a chemical formula similar to that of the invalid  compounds, allowing the LLM agent to reason more effectively and generate better informed results. 

By following the decision agent, if the answer of $\text{decide} (x_{\text{input}}, x_{\text{output}})$ is True, then we can take this optimized cathode material $x_{\text{output}}$ as the valid output.

Otherwise,  $\text{decide} (x_{\text{input}}, x_{\text{output}})$ is False, then we conduct the retrieval with the domain feedback paradigm \citep{liu2023ChatDrug}. We use the retrieval agent to retrieve the demonstration information. We search through the retrieval database, which contains lithium-based compounds downloaded from ICSD, and find a retrieved compound $x_{\text{retrieved}}$ such that it has the lowest distance ({\ie}, highest similarity) with the optimized compound $x_{\text{output}}$ yet with the correct decision result w.r.t. input compound $x_{\text{input}}$: 
\begin{equation}
x_{\text{retrieved}} = \text{min}_{x_{\text{retrieved}}} \big\{ 
    \text{d}(x_{\text{input}}, x_{\text{retrieved}})
    \land 
    \text{decide} (x_{\text{\text{input}}}, x_{\text{retrieved}}) 
\big\},
\end{equation}
where $d(\cdot)$ denotes the distance function, as will be discussed in the domain agent section below. The retrieved compounds will be used to update the prompt in the next round, as discussed in the LLM agent section above. This aims to provide improved context for the LLM agent.

\subsection{Rank Agent}
The Rank Agent aims at iteratively ranking and selecting  the most promising candidates for the next stages of the \platform{} pipeline. 

In the NMC811 optimization, 89 unique candidates remain after deduplication in Stage 5. In Stage 6, these candidates are ranked based on three prioritized criteria: total charge, preparation complexity, and voltage. Following this order, we construct a ranking tree: first selecting the top 29 candidates by total charge, then narrowing down to the top 20 based on preparation complexity, and finally selecting the top 3 based on voltage. These top 3 candidates proceed to Stage 7 for computational verification.

\subsection{Domain Agent}
In the \platform{} platform, the domain agent encapsulates various domain-specific functions to support the operations of other agents, as detailed below.

\textbf{Stage 3 Exact Match for Search Agent.} This is the exact search to find if the candidate cathode material has been reported in the Material Project:
\begin{equation}
\text{Searching}(x_{\text{output}}, D_{\text{MP}}) = \text{True} \quad \text{iff} ~~ \exists ~ x_{\text{MP}} \in D_{\text{MP}} \quad \text{s.t.}~ x_{\text{output}} = x_{\text{MP}}.
\end{equation}

\textbf{Stage 3 Range Match for Search Agent.} This is set to filter out generated  compounds that, although not identical to reported materials, contain the same chemical elements with a similar ratio of each element composition. The range search means we check if the ratio of each element from two cathode material formulas (one is the candidate cathode material $x_{\text{output}}$, and the other one is from the ICSD database $x_{\text{ICSD}}$) can match with a threshold $\tau=0.1$: 
\begin{equation}
\begin{aligned}
\text{Match}(x_{\text{output}}, D_{\text{ICSD}}) & = \text{True} \quad \text{iff} ~~ \exists ~ x_{\text{ICSD}} \in D_{\text{ICSD}}\\
& \quad \quad \text{s.t.} ~
    \frac{\text{abs}(C_{E,x_{\text{output}}} - C_{E,x_{\text{ICSD}}})} {\text{max}(C_{E,x_{\text{output}}}, C_{E,x_{\text{ICSD}}})} 
    \le \tau, \forall E \in \{x_{\text{output}}, x_{\text{ICSD}}\},
\end{aligned}
\end{equation}
where $\text{abs}(\cdot)$ is the absolute function, $E$ is the element belongs to either $x_{\text{output}}$ or $x_{\text{ICSD}}$, and $C_{E, x}$ is the count of element $E$ in molecule $x$.

\textbf{Stage 4 Theoretical Capacity Function for Decision Agent and Retrieval Agent.}
For the lithium-ion battery, we assume all lithium ions in the material participate in the electrochemical reaction. Therefore, for molecule $x$,  we consider this surrogate function for theoretical capacity calculation, which is defined as
\begin{equation}
\text{Capacity}_x = \frac{n \cdot F}{3.6 \cdot M},
\end{equation}
where $n$ is count of Li in $x$, $F=96,500 C/mol$ is Faraday’s constant and $M$ is the summation of molecular weight for $x$~\citep{li2018capacity}.

\textbf{Stage 4 Distance Function for Retrieval Agent.}
This function is designed to identify chemical formulas from our retrieval dataset of lithium-based cathode materials. The goal is to find compounds with chemical formulations similar to the invalid  compounds in terms of their chemical element composition. Chemically similar compounds with higher theoretical capacity can serve as valuable references to assist LLMs in generating valid cathode materials. To facilitate the retrieval of such similar chemical formulas, we developed a distance function that quantifies the similarity between two molecules, $a$ and $b$, based on element counts.

To construct the function, we define the element counts across seven levels within the material formulas, along with their corresponding weights.
\begin{itemize}[noitemsep,topsep=0pt]
    \item $C_1,x$ is defined as the count of \#(Li) in molecule $x$. The primary focus of our retrieval dataset is lithium-based cathode materials. Therefore, Li is a fundamental element, and nearly all retrieved compounds inherently contain lithium. As a result, the weighting coefficient associated with $C_1,x$ ($\alpha$) should be set relatively low to avoid excessive penalization when comparing lithium-containing compounds.
    \item $C_2,x$ is defined as the count of \#(Mn, Co, Ni) in molecule $x$. Mn, Co, and Ni are the transition metals cations in NMC811. Since these transition metals are integral to the NMC811 structure, any significant variation in their counts between compounds can indicate a substantial difference in material properties. Thus, its weighting coefficient ($\beta$) should be set relatively high to reflect their critical role.
    \item $C_3,x$ is defined as the count of \#(Fe, Cu, Zn, V, Cr, Ti, Mo) in molecule $x$. These elements represent typical cations, all of which are alternative transition metals commonly used in cathode materials. Distinguishing between these cations is important for accurately categorizing and retrieving relevant cathode materials. Therefore, its weighting coefficient ($\gamma$) should be assigned a moderate value, similar to $C_5,x$.
    \item $C_4,x$ is defined as the count of \#(O, P, F, S, Cl, Br, I) in molecule $x$. The chemical and electrochemical properties of cathode materials are heavily influenced by their anionic composition, especially in terms of oxygen content and other functional groups, so the weighting coefficient ($\delta$) should be set as a significantly high value to effectively differentiate between compounds.
    \item $C_5,x$ is defined as the count of \#(Mg, Al, Si, B, Zr, C, Be, Ca, Na, K, Sn, Sr) in molecule $x$. These elements are other typical cations, most of which belong to the alkali metals group, alkaline earth metals group, boron group, and carbon group. Due to their frequent use in cathode compounds, distinguishing these cations is essential for precise classification of cathode compounds. Therefore, the weighting coefficient ($\epsilon$) should be set to a moderate value, comparable to that of $C_3,x$.
    \item $C_6,x$ is defined as the count of any other elements in molecule $x$. These elements are less common in cathode materials, so the weighting coefficient ($\zeta$) should be set to the lowest value.
    \item $C_{\text{species}},x$ is defined as the count of element types in molecule $x$. Diverse elemental composition can significantly impact the electrochemical properties and the suitability of cathode materials. Therefore, its weighting coefficient ($\eta$) should be significantly high to penalize large differences in elemental diversity.
\end{itemize}
After testing various combinations of different weights for the seven levels, the weighting coefficients ($\alpha$, $\beta$, $\gamma$, $\delta$, $\epsilon$, $\zeta$, $\eta$) that produce stable and reasonable retrieval results are set to the values of 3, 7, 5, 10, 5, 1, and 10, respectively. The final distance between molecule $a$ and molecule $b$ is defined as:
\begin{equation}
\begin{aligned}
\text{d}(a,b) = 
& ~~ 3* \big|C_{1,a} - C_{1,b}\big| + 7* \big|C_{2,a} - C_{2,b}\big| + 5* \big|C_{3,a} - C_{3,b}\big| + 10* \big|C_{4,a} - C_{4,b}\big| + \\
& ~~ 5*\big|C_{5,a} - C_{5,b}\big| + 1*\big|C_{6,a} - C_{6,b}\big| + 10 * \big|C_{\text{species},a} - C_{\text{species},b}\big|.
\end{aligned}
\end{equation}
The smaller the value of $d$, the closer the two formulas become.

\textbf{Stage 5 Range Match for Deduplication.}
We apply range matching, as discussed previously, to eliminate duplicate hypotheses. After Phase 1 exploration, we obtain 100 candidate materials. We then scan through the list, and for any pair of candidates $x_i$ and $x_j$ with indices $1 \le i < j \le 100$, if they are considered a match, {\ie},
\begin{equation}
\frac{\text{abs}(C_{E,x_i} - C_{E,x_j})} {\text{max}(C_{E,x_i}, C_{E,x_i})} \le \tau, ~~ \forall E \in \{x_i, x_j\},
\end{equation}
then we remove the duplicated candidate $x_j$.

\textbf{Stage 6 Total Charge Calculation for Ranking.} This total charge calculation is designed to ensure consistency and comparability in theoretical capacity and voltage between NMC811 and the generated candidates. The total charge of the native compound is zero, and the same elements in various compounds may exhibit different valence states. The valence states of the constituent elements determine various properties of the compound, such as theoretical capacity and voltage.

Therefore, to ensure a fair comparison between the theoretical capacities and voltages of the generated compounds and NMC811, the valence states of the elements present in the generated compounds that also exist in NMC811 should be identical to those in NMC811. Specifically, the valence states in NMC811 are assigned as integers, and achieve overall charge neutrality, with Li, Mn, Co, Ni, and O set to +1, +3, +3, +3, and –2, respectively. For elements not present in NMC811, their valence states are set to the highest oxidation state achievable under non-acidic salt conditions to reflect the theoretical upper bound of redox activity each element can contribute. Notably, the valence values of these elements are not precise formal charges derived from DFT calculations, but instead represent approximations based on oxidation state heuristics.

We estimate the total charge of each molecule $x$ based on oxidation states as:
\begin{equation}
\text{Total Charge}_x = C_{E, x} * V_E, ~~ \forall E \in \{x\},
\end{equation}
where $V_E$ is the valence values of element $E$. We list the valence values in \Cref{tab:valence_values_for_total_charge}.

After calculating the total charge for each compound, we use this metric as the first criterion for ranking the generated cathode materials. While 11 compounds have a total charge of exactly zero, we account for the variability in oxidation states—particularly among transition metals—and the possible valence deviations during experimental synthesis. To reflect this, we prioritize compounds whose total charges are closer to zero. Specifically, we rank all candidates based on the absolute value of their total charge and select the top 29 compounds with charges closest to zero for subsequent evaluation of preparation complexity.

\begin{table}[h]
\centering
\caption{\small T\textbf{he valence values of elements.}} \label{tab:valence_values_for_total_charge}
\vspace{-2ex}
\begin{tabular}{lc | lc | lc}
\toprule
Elements & Valence Values & Elements & Valence Values & Elements & Valence Values\\
\midrule
C      & 4 & Si     &4 & Ge     &4\\
Sn     & 4 & Pb     &4 & Be     &2\\  
Mg     &2 & Ca     &2 & Sr     &2\\
Ba     &2 & Sc     &3 & Ti     &4\\
V      &3 & Cr     &3 & Mn     &3\\
Fe     &3 & Co     &3 & Ni     &3\\
Cu     &2 & Zn     &2 & Mo     &6\\
Zr     &4 & Y      &3 & Li     &1\\
O      &-2 & Na     &1 & K      &1\\
B      &3 & Al     &3 & Ga     &3\\
\bottomrule
\end{tabular}
\end{table}

\section*{Data Availability}
The data is available on this \href{https://huggingface.co/datasets/chao1224/ChatBattery}{HuggingFace link}.

\section*{Code Availability}
The code is available on this \href{https://github.com/chao1224/ChatBattery}{GitHub repository link}. Notably, we use the Flask package to provide a website as the user interface system for stages 1 to 4, which is more convenient for domain scientists. Please check the repository for more details.

\section*{Acknowledgment}
This project was funded by a National Research Council Canada (NRC) collaborative R\&D grant (AI4D-core-132). Y.B. acknowledges funding from NRC AI4D, CIFAR and CIFAR AI Chair.  H.L. acknowledges the Postdoctoral Fellowship Program (Grant No. PC2022020) and the use of the University of Oxford Advanced Research Computing (ARC) facility for resources in carrying out this work (10.5281/zenodo.22558). H.L. and H.X. gratefully thank University of Oxford for providing access to the Inorganic Crystal Structure Database (ICSD). 

\section*{Author Contribution Statement}
The discussion of ideas and paper writing was contributed to by everyone. The code implementation and platform design were contributed by S.L. and Y.A. The experiment running for stages 1 to 8, hyperparameter selection, and the domain insights were contributed by H.X., H.L., and H.G. Y.B. and H.G. led and managed the overall project.
\clearpage

{
\renewcommand*{\bibfont}{\small}
\printbibliography[title={References}]
}

\newpage
\appendix

\renewcommand{\thetable}{S\arabic{table}}
\captionsetup[table]{name=Supplementary Table}
\setcounter{table}{0}

\renewcommand{\thefigure}{S\arabic{figure}}
\captionsetup[figure]{name=Supplementary Figure}
\setcounter{figure}{0}

\section{Experimental Details for NMC811 Optimization}

The following tables summarize key stages in the exploitation phase of NMC811 optimization.

\begin{itemize}[noitemsep,topsep=0pt]
    \item \Cref{tab:SI_01} presents the 20 cathode materials after the first cycle and 100 cathode materials after the second cycle.
    \item \Cref{tab:SI_02} presents the list of 11 duplicate cathode materials with range match.
    \item \Cref{tab:SI_03} presents the list of 29 cathode materials ranked by the absolute total charge.
    \item \Cref{tab:SI_04} presents the list of 20 cathode materials ranked by preparation complexity.
    \item \Cref{tab:SI_05} presents the list of 3 cathode materials ranked by the LLM-surrogated voltage.
\end{itemize}

\begin{longtable}{ l l }
\caption{\small \textbf{The list of 20 optimized cathode materials after the first cycle and 100 optimized cathode materials after the second cycle.}} \label{tab:SI_01}
\\

\toprule
Cathodes  List of 1st Cycle of Opt. & Cathodes  List of 2nd Cycle of Opt. \\
\midrule
\endfirsthead

\toprule
Cathodes  List of 1st Cycle of Opt. & Cathodes  List of 2nd Cycle of Opt. \\
\midrule

\endhead

LiNi0.3Mn0.1Co0.1Si0.5O2  & LiNi0.23Mn0.11Co0.11Si0.36Mg0.19O2 \\
                          & LiNi0.24Mn0.11Co0.11Si0.38Mg0.16O2 \\
                          & LiNi0.26Mn0.11Co0.11Si0.41Mg0.11O2 \\
                          & LiNi0.23Mn0.13Co0.13Si0.40Mg0.11O2 \\
                          & LiNi0.25Mn0.1Co0.1Si0.45Mg0.1O2    \\
\cline{1-2} 
LiNi0.5Mn0.1Co0.1Mg0.2O2 & LiNi0.25Mn0.1Co0.1Mg0.15V0.3O2 \\
                         & LiNi0.2Mn0.1Co0.1Mg0.2V0.3O2 \\
                         & LiNi0.4Mn0.1Co0.1Mg0.15Al0.25O2 \\
                         & LiNi0.35Mn0.15Fe0.1Mg0.2O2 \\
                         & LiNi0.4Mn0.15Co0.05Mg0.15Si0.25O2    \\
\cline{1-2} 
LiNi0.4Mn0.1Co0.1Ca0.4O2 & LiNi0.25Mn0.2Co0.1Ca0.3Al0.15O2 \\
                         & LiNi0.3Mn0.1Co0.1Ca0.3Al0.2O2 \\
                         & LiNi0.25Mn0.15Co0.2Si0.4O2 \\
                         &  LiNi0.35Mn0.15Co0.15Al0.35O2 \\
                         & LiNi0.3Mn0.1Co0.1Ca0.3Si0.2O2    \\
\cline{1-2} 
LiNi0.65Mn0.15Co0.1Ca0.1O2 & LiNi0.6Mn0.15Co0.1Ca0.1Ti0.05O2 \\
                           & LiNi0.55Mn0.15Co0.1Ca0.1Ti0.05Si0.05O2 \\
                        & LiNi0.5Mn0.15Co0.1Ca0.1Ti0.05Si0.05C0.05O2 \\
                        & LiNi0.45Mn0.15Co0.1Ca0.1Ti0.05Si0.05C0.1O2 \\
                        & LiNi0.4Mn0.15Co0.1Ca0.1Ti0.05Si0.05C0.2O2  \\
\cline{1-2} 
LiNi0.75Mn0.05Co0.05Si0.15O2 & LiNi0.4Mn0.05Co0.05Si0.15Ca0.35O2 \\
                             & LiNi0.5Mn0.05Co0.05Si0.15Ca0.25O2 \\
                             & LiNi0.6Mn0.05Co0.05Si0.15Ca0.15O2 \\
                             & LiNi0.7Mn0.05Co0.05Si0.1Mg0.1O2 \\
                             & LiNi0.65Mn0.05Co0.05Si0.15Ca0.1O2    \\
\cline{1-2}
LiNi0.65Mn0.1Co0.1Ca0.15O2 & LiNi0.4Mn0.1Co0.1Ca0.2Sc0.2O2 \\
                           & LiNi0.6Mn0.1Co0.1Ca0.1Ti0.1O2 \\
                           & LiNi0.55Mn0.1Co0.1Ca0.15Al0.1O2 \\
                           & LiNi0.6Mn0.1Co0.1Ca0.1Si0.1O2 \\
                           & LiNi0.45Mn0.1Co0.1Ca0.2B0.15O2    \\
\cline{1-2} 
LiNi0.7Mn0.1Co0.1Si0.1O2 & LiNi0.5Mn0.1Co0.1Si0.1Ca0.1O2 \\
                         & LiNi0.6Mn0.1Co0.1Si0.1Ca0.1O2 \\
                         & LiNi0.6Mn0.1Co0.1Si0.1Ti0.1O2 \\
                         & LiNi0.6Mn0.1Co0.1Si0.1Mg0.1O2 \\
                         & LiNi0.65Mn0.1Co0.1Si0.1Ca0.05O2    \\
\hline
LiNi0.3Mn0.1Co0.1Si0.3O2  & Li1.22Mn0.38Co0.14Ni0.26Ca0.02O2 \\
                          & Li1.22Mn0.38Co0.14Ni0.26Sr0.02O2 \\
                          & Li1.22Mn0.38Co0.14Ni0.26Zn0.02O2 \\
                          & Li1.22Mn0.38Co0.14Ni0.26Ba0.02O2 \\
                          & Li1.22Mn0.38Co0.14Ni0.26Mg0.02O2    \\
\cline{1-2} 
LiNi0.5Mn0.1Co0.1Al0.1O2 & LiNi0.4Mn0.1Co0.1Al0.1Ti0.1O2 \\
                         & LiNi0.45Mn0.1Co0.1Al0.1Ca0.05O2 \\
                         & LiNi0.4Mn0.1Co0.1Al0.1Mg0.1O2 \\
                         & LiNi0.35Mn0.1Co0.1Al0.1Si0.15O2 \\
                         & LiNi0.3Mn0.1Co0.1Al0.1B0.2O2    \\
\cline{1-2} 
LiNi0.4Mn0.1Co0.1Mg0.2O2 & LiNi0.3Mn0.1Co0.1Mg0.2Al0.1O2 \\
                         & LiNi0.3Mn0.1Co0.1Mg0.2Ti0.1O2 \\
                         & LiNi0.3Mn0.1Co0.1Mg0.2Fe0.1O2 \\
                         & LiNi0.25Mn0.1Co0.1Mg0.2Cr0.15O2 \\
                         & LiNi0.3Mn0.1Co0.1Mg0.2Si0.1O2    \\
\cline{1-2} 
LiNi0.7Mn0.1Co0.1Ti0.1O2 & LiNi0.6Mn0.1Co0.1Ti0.1Fe0.1O2 \\
                         & LiNi0.5Mn0.1Co0.1Ti0.1Fe0.1Al0.1O2 \\
                         & LiNi0.4Mn0.1Co0.1Ti0.1Fe0.1Al0.1Si0.1O2 \\
            & LiNi0.3Mn0.1Co0.1Ti0.1Fe0.1Al0.1Si0.1Mg0.1O2 \\
            & LiNi0.2Mn0.1Co0.1Ti0.1Fe0.1Al0.1Si0.1Mg0.1Zn0.1O2 \\
\cline{1-2}
LiNi0.7Mn0.1Co0.1Ti0.1O2 & LiNi0.5Mn0.1Co0.1Ti0.1Fe0.1Ca0.1O2 \\
                         & LiNi0.5Mn0.1Co0.1Ti0.1Fe0.1Mg0.1O2 \\
                         & LiNi0.4Mn0.1Co0.1Ti0.1Fe0.1Si0.05O2 \\
                         & LiNi0.3Mn0.1Co0.1Ti0.1Fe0.1B0.05O2 \\
                         & LiNi0.2Mn0.1Co0.1Ti0.1Fe0.1C0.05O2    \\
\cline{1-2} 
LiNi0.7Mn0.1Co0.1Ti0.1O2 & LiNi0.6Mn0.1Co0.1Ti0.1Mg0.1O2 \\
                         & LiNi0.65Mn0.1Co0.1Ti0.1Al0.05O2 \\
                         & LiNi0.55Mn0.1Co0.1Ti0.1Ca0.15O2 \\
                         & LiNi0.6Mn0.1Co0.1Ti0.1Si0.1O2 \\
                         & LiNi0.5Mn0.1Co0.1Ti0.1B0.3O2    \\
\cline{1-2} 
LiNi0.7Mn0.1Co0.1Ti0.1O2 & LiNi0.6Mn0.1Co0.1Ti0.1Mg0.1O2 \\
                         & LiNi0.65Mn0.1Co0.1Ti0.1Al0.05O2 \\
                         & LiNi0.55Mn0.1Co0.1Ti0.1Ca0.15O2 \\
                         & LiNi0.6Mn0.1Co0.1Ti0.1Si0.1O2 \\
                         & LiNi0.5Mn0.1Co0.1Ti0.1B0.3O2    \\
\cline{1-2} 
LiNi0.6Mn0.1Co0.1Zr0.1O2 & LiNi0.15Mn0.1Co0.1Zr0.1Ti0.55O2 \\
                         & LiNi0.25Mn0.15Co0.15Zr0.1Ca0.35O2 \\
                         & LiNi0.4Mn0.15Co0.15Zr0.1Al0.2O2 \\
                         & LiNi0.4Mn0.1Co0.1Zr0.1Al0.3O2 \\
                         & LiNi0.35Mn0.15Co0.15Zr0.1Si0.25O2    \\
\cline{1-2} 
LiNi0.6Mn0.1Co0.1Zr0.1O2 & LiNi0.1Mn0.1Co0.1Zr0.1Ti0.6O2 \\
                         & LiNi0.4Mn0.1Co0.1Zr0.1Mg0.3O2 \\
                         & LiNi0.3Mn0.1Co0.1Zr0.1Ca0.4O2 \\
                         & LiNi0.4Mn0.1Co0.1Zr0.1Si0.3O2 \\
                         & LiNi0.5Mn0.1Co0.1Zr0.1Al0.2O2     \\
\cline{1-2} 
LiNi0.6Mn0.1Co0.1Al0.1O2 & LiNi0.25Mn0.2Co0.2Al0.1Si0.25O2 \\
                         & LiNi0.3Mn0.1Co0.1Al0.1Ca0.3O2 \\
                         & LiNi0.4Mn0.1Co0.1Al0.1Si0.3O2 \\
                         & LiNi0.35Mn0.15Co0.15Al0.1Ca0.2O2 \\
                         & LiNi0.35Mn0.15Co0.15Al0.1Si0.25O2    \\
\cline{1-2} 
LiNi0.6Mn0.1Co0.1Al0.1O2 & LiNi0.3Mn0.1Co0.1Al0.1V0.1Sr0.1O2 \\
                         &  LiNi0.4Mn0.1Co0.1Al0.1Ti0.1C0.1O2 \\
                         & LiNi0.4Mn0.1Co0.1Al0.1Fe0.1Mg0.1O2 \\
                         &  LiNi0.3Mn0.1Co0.1Al0.1Ti0.1C0.1O2 \\
                         & LiNi0.4Mn0.1Co0.1Al0.1Ti0.1C0.2O2   \\
\cline{1-2} 
LiNi0.8Mn0.05Co0.05Mg0.05O2 & LiNi0.7Mn0.03Co0.03Mg0.03Sn0.01O2 \\
                            & LiNi0.7Mn0.05Co0.05Mg0.05Ti0.05O2 \\
                            & LiNi0.75Mn0.03Co0.03Mg0.03Al0.06O2 \\
                            & LiNi0.6Mn0.1Co0.1Mg0.1Si0.1O2 \\
                            & LiNi0.65Mn0.07Co0.07Mg0.07Zr0.04O2     \\
\cline{1-2} 
LiNi0.8Mn0.05Co0.05Mg0.05O2 & LiNi0.75Mn0.05Co0.05Mg0.05Y0.03O2 \\
                            & LiNi0.7Mn0.05Co0.05Mg0.05Ti0.05O2 \\
                            & LiNi0.75Mn0.05Co0.05Mg0.05Al0.05O2 \\
                            & LiNi0.6Mn0.1Co0.1Mg0.1Si0.1O2 \\
                            & LiNi0.65Mn0.1Co0.1Mg0.1B0.05O2    \\
\bottomrule
\end{longtable}

\clearpage
\vspace*{-1cm}
\begin{table}[ht]
\centering
\caption{\small \textbf{ List of 11 duplicated cathode materials after the range match.}} \label{tab:SI_02}
\begin{tabular}{l}
\toprule
Cathode Materials List\\
\toprule
LiNi0.25Mn0.1Co0.1Si0.45Mg0.1O2\\
LiNi0.6Mn0.1Co0.1Si0.1Mg0.1O2\\
LiNi0.6Mn0.1Co0.1Ca0.1Si0.1O2\\
LiNi0.6Mn0.1Co0.1Ti0.1Si0.1O2\\
LiNi0.7Mn0.05Co0.05Mg0.05Ti0.05O2\\
LiNi0.6Mn0.1Co0.1Mg0.1Si0.1O2\\
LiNi0.6Mn0.1Co0.1Ti0.1Mg0.1O2\\
LiNi0.65Mn0.1Co0.1Ti0.1Al0.05O2\\
LiNi0.55Mn0.1Co0.1Ti0.1Ca0.15O2\\
LiNi0.6Mn0.1Co0.1Ti0.1Si0.1O2\\
LiNi0.5Mn0.1Co0.1Ti0.1B0.3O2\\
\bottomrule
\end{tabular}
\end{table}

\clearpage
\begin{table}[ht]
\centering
\caption{\small \textbf{List of 29 cathode materials ranked by the absolute total charges, ranging from  –0.1e to 0.1e.}} \label{tab:SI_03}
\begin{tabular}{lr}
\toprule
Cathode Materials List & Total Charge (e) \\
\midrule
LiNi0.35Mn0.15Co0.15Al0.35O2               &       	0.0 \\
LiNi0.5Mn0.1Co0.1Ti0.1Fe0.1Ca0.1O2         &       	0.0 \\
LiNi0.5Mn0.1Co0.1Ti0.1Fe0.1Mg0.1O2         &       	0.0 \\
LiNi0.6Mn0.1Co0.1Mg0.1Si0.1O2              &       	0.0 \\
LiNi0.6Mn0.1Co0.1Si0.1Ca0.1O2              &       	0.0 \\
LiNi0.6Mn0.1Co0.1Ca0.1Ti0.1O2              &       	0.0 \\
LiNi0.6Mn0.1Co0.1Ti0.1Mg0.1O2              &       	0.0 \\
LiNi0.2Mn0.1Co0.1Ti0.1Fe0.1Al0.1Si0.1Mg0.1Zn0.1O2 &	4.440892098500626e-16 \\
LiNi0.6Mn0.05Co0.05Si0.15Ca0.15O2                 &	-4.440892098500626e-16 \\
LiNi0.7Mn0.05Co0.05Si0.1Mg0.1O2                   &	-4.440892098500626e-16 \\
LiNi0.55Mn0.15Co0.1Ca0.1Ti0.05Si0.05O2            &	6.661338147750939e-16 \\
LiNi0.6Mn0.15Co0.1Ca0.1Ti0.05O2                   &	-0.04999999999999982 \\
LiNi0.65Mn0.1Co0.1Si0.1Ca0.05O2                   &	 0.04999999999999982 \\
LiNi0.55Mn0.1Co0.1Ti0.1Ca0.15O2                   & -0.050000000000000044 \\
LiNi0.65Mn0.05Co0.05Si0.15Ca0.1O2                 & 	0.050000000000000266 \\
LiNi0.5Mn0.15Co0.1Ca0.1Ti0.05Si0.05C0.05O2        & 	 0.05000000000000071 \\
LiNi0.4Mn0.1Co0.1Zr0.1Al0.3O2                     & 	 0.09999999999999964 \\
LiNi0.6Mn0.1Co0.1Ti0.1Fe0.1O2                     & 	 0.09999999999999964 \\
LiNi0.5Mn0.1Co0.1Ti0.1Fe0.1Al0.1O2                & 	0.09999999999999964 \\
LiNi0.4Mn0.15Co0.05Mg0.15Si0.25O2                 & 	0.09999999999999964 \\
LiNi0.65Mn0.1Co0.1Ti0.1Al0.05O2                   & 	0.09999999999999964 \\
LiNi0.65Mn0.1Co0.1Mg0.1B0.05O2                    & 	-0.09999999999999964 \\
LiNi0.5Mn0.1Co0.1Zr0.1Al0.2O2                     & 	0.09999999999999964 \\
LiNi0.3Mn0.1Co0.1Ca0.3Si0.2O2                     & 	-0.09999999999999987 \\
LiNi0.5Mn0.05Co0.05Si0.15Ca0.25O2                 & 	-0.10000000000000009 \\
LiNi0.3Mn0.1Co0.1Ti0.1Fe0.1Al0.1Si0.1Mg0.1O2      & 	0.10000000000000031 \\
LiNi0.4Mn0.15Co0.15Zr0.1Al0.2O2                   & 	0.10000000000000053 \\
LiNi0.4Mn0.1Co0.1Al0.1Ti0.1C0.1O2                 & 	-0.10000000000000053 \\
LiNi0.45Mn0.15Co0.1Ca0.1Ti0.05Si0.05C0.1O2        & 	0.10000000000000098 \\
\bottomrule
\end{tabular}
\end{table}

\newpage
\clearpage
\begin{table}[ht]
\centering
\caption{\small \textbf{List of 20 cathode materials ranked by preparation complexity.}} \label{tab:SI_04}
\begin{tabular}{l r}
\toprule
Cathode Materials List & Preparation Complexity \\
\toprule
LiNi0.35Mn0.15Co0.15Al0.35O2                      &6\\
LiNi0.6Mn0.1Co0.1Mg0.1Si0.1O2                     &7\\
LiNi0.6Mn0.1Co0.1Si0.1Ca0.1O2                     &7\\
LiNi0.6Mn0.1Co0.1Ca0.1Ti0.1O2                     &7\\
LiNi0.6Mn0.1Co0.1Ti0.1Mg0.1O2                     &7\\
LiNi0.6Mn0.05Co0.05Si0.15Ca0.15O2                 &7\\
LiNi0.7Mn0.05Co0.05Si0.1Mg0.1O2                   &7\\
LiNi0.6Mn0.15Co0.1Ca0.1Ti0.05O2                   &7\\
LiNi0.65Mn0.1Co0.1Si0.1Ca0.05O2                   &7\\
LiNi0.55Mn0.1Co0.1Ti0.1Ca0.15O2                   &7\\
LiNi0.65Mn0.05Co0.05Si0.15Ca0.1O2                 &7\\
LiNi0.4Mn0.1Co0.1Zr0.1Al0.3O2                     &7\\
LiNi0.6Mn0.1Co0.1Ti0.1Fe0.1O2                     &7\\
LiNi0.4Mn0.15Co0.05Mg0.15Si0.25O2                 &7\\
LiNi0.65Mn0.1Co0.1Ti0.1Al0.05O2                   &7\\
LiNi0.65Mn0.1Co0.1Mg0.1B0.05O2                    &7\\
LiNi0.5Mn0.1Co0.1Zr0.1Al0.2O2                     &7\\
LiNi0.3Mn0.1Co0.1Ca0.3Si0.2O2                     &7\\
LiNi0.5Mn0.05Co0.05Si0.15Ca0.25O2                 &7\\
LiNi0.4Mn0.15Co0.15Zr0.1Al0.2O2                   &7\\
\bottomrule
\end{tabular}
\end{table}

\begin{table}[h]
\centering
\caption{\small \textbf{List of 3 cathode materials ranked by voltage surrogate.}} \label{tab:SI_05}
\begin{tabular}{l r}
\toprule
Cathode Materials List & Voltage Ranking Index \\
\toprule
LiNi0.7Mn0.05Co0.05Si0.1Mg0.1O2 & 1\\
LiNi0.65Mn0.1Co0.1Mg0.1B0.05O2 & 2\\
LiNi0.65Mn0.1Co0.1Si0.1Ca0.05O2 & 3\\
\bottomrule
\end{tabular}
\end{table}

\clearpage
\newpage
\section{The Characterization of Li-rich-NMC-SiMg} 
\label{sec:wet_lab_LirNMCSiMg}

We chose NMC-SiMg for further optimization based on domain knowledge and prior experimental insights. The NMC-SiMg cathode was further optimized by increasing the lithium content within the crystal lattice, yielding a Li-rich variant, Li-rich-NMC-SiMg, designed to achieve higher capacity. This material was motivated in part by previous studies on Li-rich layered oxides, such as Li\textsubscript{1.2}Ni\textsubscript{0.13}Mn\textsubscript{0.54}Co\textsubscript{0.13}O\textsubscript{2}~\citep{song2024visualization}. Li-rich layered oxides revealed the formation of molecular oxygen (O\textsubscript{2}) trapped within transition metal vacancies, a phenomenon associated with structural instability. The incorporation of Si and Mg is hypothesized to mitigate this issue by forming strong bonds with oxygen, thereby stabilizing the Li-rich framework. Consequently, Li-rich-NMC-SiMg was synthesized by increasing the lithium stoichiometry to 1.2 using the same synthetic methodology of the NMC811-derived samples. 

Li-rich-NMC-SiMg exhibits enhanced electrochemical performance, achieving an initial discharge capacity over 200 mAh/g and a reversible capacity of 181 mAh/g by the third cycle, with an Initial Coulombic Efficiency (ICE) of up to 83\%. Compared to NMC811, this represents a 34\% improvement over NMC811 under identical conditions, and surpasses conventional cathode materials with typical capacities of 120–150 mAh/g. 

\begin{figure}[ht]
    \centering
    \includegraphics[width=\linewidth]{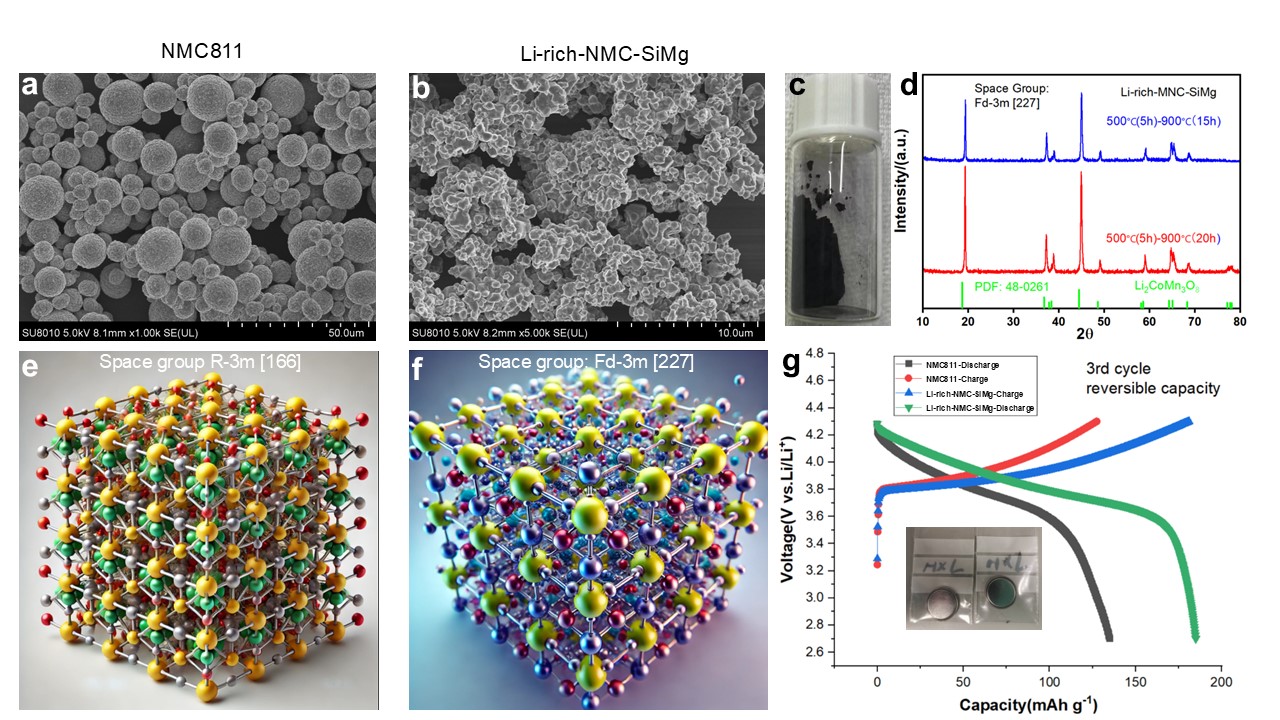} 
    \caption{\small \textbf{(a) The SEM image of NMC811 and (b) Li-rich-NMC-SiMg; (c) the optical image of final product NMC-SiMg; (d) the XRD patterns of Li-rich-NMC-SiMg produced at different temperatures; (e) the visual images of NMC (Li (yellow spheres), transition metals (green octahedra for Ni, Mn, Co), and oxygen (red spheres)) with space group R-3m and (f) the Li-rich-NMC-SiMg (the image highlights lithium (yellow spheres), transition metals (Ni, Mn, Co in green octahedra), silicon (blue spheres), magnesium (purple spheres), and oxygen (red spheres) within a repeating cubic framework) with space group Fd-3m; (g) the reversible capacities (3rd cycle) of NMC and Li-rich-NMC-SiMg.}}
    \label{fig:SF1}
\end{figure}

The SEM images of NMC811 and Li-rich-NMC-SiMg are shown in~\Cref{fig:SF1}a and b, respectively. NMC811 exhibits a typical spherical morphology, which is commonly found in current commercial cathode materials for lithium-ion batteries~\citep{wijareni2022morphology}. In contrast, Li-rich-NMC-SiMg displays uniform and interconnected primary particles, with no obvious formation of secondary particles. This unique morphology of Li-rich-NMC-SiMg presents a trade-off: while it enhances lithium-ion transport and capacity, it also affects the formation of the solid electrolyte interphase (SEI) and lowers the initial coulombic efficiency (ICE). Further optimization of the Li-rich-NMC-SiMg morphology could help balance the improvement in capacity with better ICE performance~\citep{ahsan2024recent, kang2023progress}. The optical image of Li-rich-NMC-SiMg, shown in~\Cref{fig:SF1}c, reveals a typical fine powder state, suggesting that it could be used to form an ideal ink with appropriate viscosity and rheological properties for subsequent coating processes, which are crucial for battery fabrication. The XRD patterns of Li-rich-NMC-SiMg, presented in~\Cref{fig:SF1}d, show sharp peaks across various synthesis temperatures, indicating the material's tendency to form a crystalline structure. The space group is Fd-3m [227], which is similar to that of Li-rich materials such as Li\textsubscript{2}CoMn\textsubscript{3}O\textsubscript{8} (PDF No. 48-0261)~\citep{ghiyasiyan2018effect,meshram2022solvent}. In comparison, NMC811 exhibits a layered rhombohedral structure with lithium ions located within the layers (space group: R-3m [166]), as shown in the rendered image in~\Cref{fig:SF1}e. On the other hand, Li-rich-NMC-SiMg adopts a cubic layered arrangement with lithium (yellow spheres) occupying the corners (space group: Fd-3m [227]), as illustrated in~\Cref{fig:SF1}f. The differences in crystal structures indicate a transition from a layered structure to a cubic one with the addition of Si and Mg elements in Li-rich-NMC-SiMg. To evaluate the electrochemical performance, NMC811 and Li-rich-NMC-SiMg were assembled into standard coin cells (2032 type). For a more meaningful comparison of reversible capacity, the charge-discharge profiles at the third cycle for both NMC811 and Li-rich-NMC-SiMg are shown in~\Cref{fig:SF1}g. Both materials exhibit a Coulombic efficiency (CE) approaching 100\%, indicating fully reversible capacity. Notably, Li-rich-NMC-SiMg demonstrates a capacity of 181 mAh/g, which is 34\% higher than that of NMC811 (135 mAh/g), confirming that the addition of Si and Mg elements improves the electrochemical performance of lithium-ion batteries as a cathode material, driven by the changes in morphology and crystal structure. The enhanced capacity arises from the increased availability of active lithium ions, as the crystal structure of Li-rich-NMC-SiMg resembles that of Li-rich cathode materials. Furthermore, the addition of Si and Mg enhances conductivity, lithium-ion diffusion, and structural integrity, while also reducing oxygen release and degradation. 

\begin{figure}[!t]
    \centering
    \includegraphics[width=0.65\linewidth]{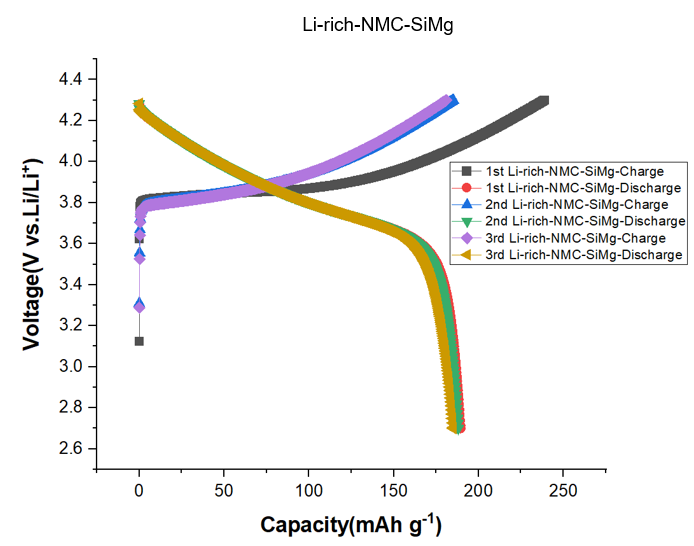}
    \caption{\small \textbf{The charge-discharge profiles of Li-rich-NMC-SiMg electrode for the first 3 cycles.}}
    \label{fig:SF2}
\end{figure}

\begin{figure}[h]
    \centering
    \includegraphics[width=0.65\linewidth]{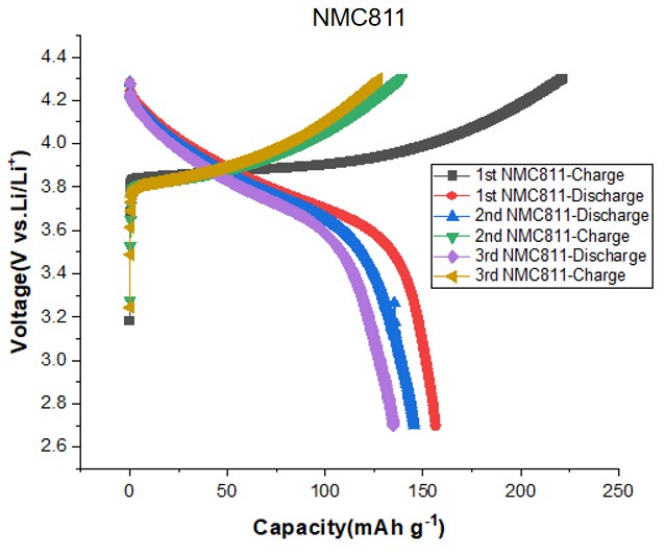}
    \caption{\small \textbf{The charge-discharge profiles of NMC811 electrode for the first 3 cycles.}}
    \label{fig:SF3}
\end{figure}

\begin{figure}[!t]
    \centering
    \includegraphics[width=0.65\linewidth]{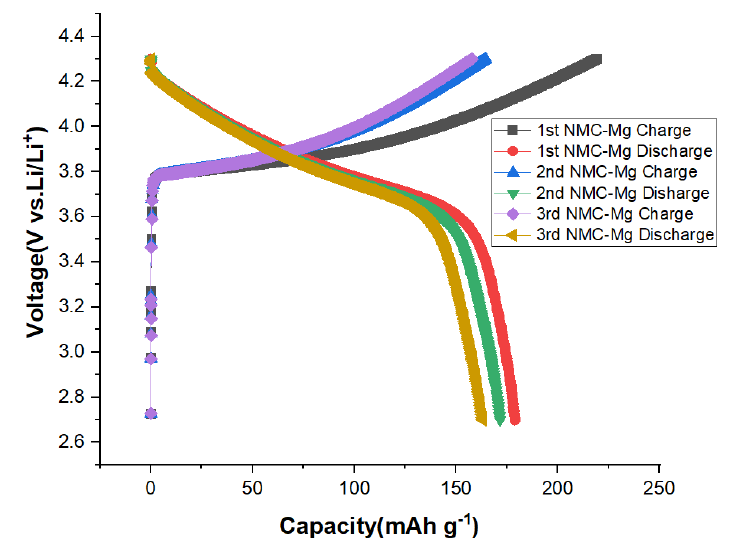} 
    \caption{\small \textbf{The charge-discharge profiles of NMC-Mg electrode for the first 3 cycles.}}
    \label{fig:SF4}
\end{figure}

In addition, the charge-discharge profiles for the first three cycles of the Li-rich-NMC-SiMg, NMC811, and NMC-Mg electrodes are presented in~\Cref{fig:SF2,fig:SF3,fig:SF4}, respectively. The results demonstrate that the Li-rich-NMC-SiMg electrode exhibits significantly higher capacity and enhanced stability compared to the NMC811 and NMC-Mg electrodes. These roots for performance improvements are consistent with the results from previous \platform{} reasoning, which is expected to predict new materials regarding electrodes, electrolytes, and additives in extensive battery systems~\citep{li2019ultra,li2020altering,li2023ampere}.

\begin{figure}[!t]
    \centering
    \includegraphics[width=\linewidth]{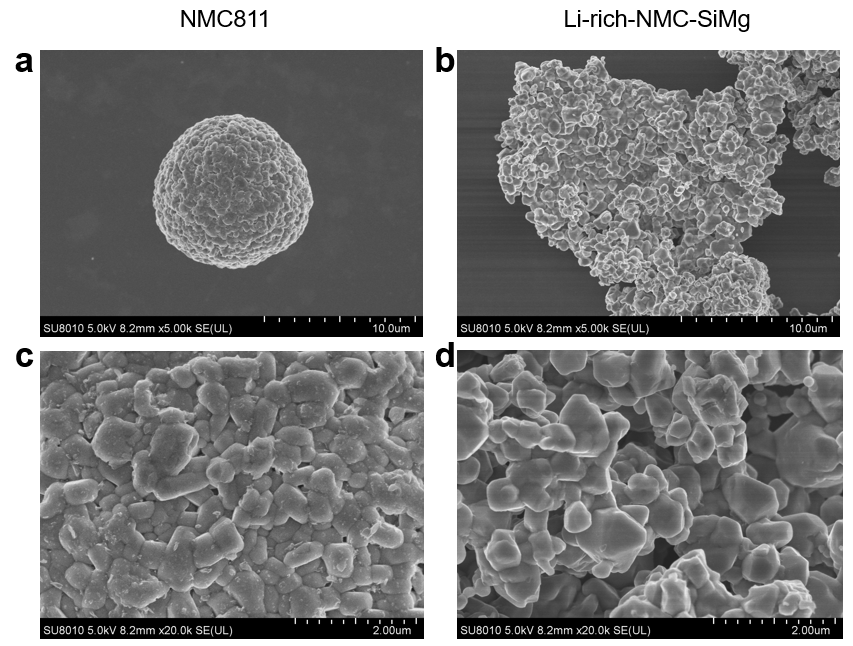} 
    \caption{\small \textbf{The SEM images of (a) low resolution for NMC811, and (b) Li-rich-NMC-SiMg, and high resolution for (c) NMC811 and (d) Li-rich-NMC-SiMg.}}
    \label{fig:SF5}
\end{figure}

SEM analysis further highlights the nanoscale microstructural differences between NMC811 and Li-rich-NMC-SiMg. As shown in~\Cref{fig:SF5}, NMC811 exhibits a spherical morphology composed of secondary particles, with particle sizes ranging from 5 to 15 $\mu$m (\Cref{fig:SF5}a). These spherical particles are densely packed with numerous layered primary nanocrystallites (\Cref{fig:SF5}c). This structure contributes to improved tap density and structural stability, while also facilitating uniform lithium-ion diffusion within the electrode. In contrast, Li-rich-NMC-SiMg displays a prominent flake-like layered structure composed of secondary particles (\Cref{fig:SF5}b), and its high-resolution image reveals layered primary particles, similar to those observed in NMC811 (\Cref{fig:SF5}d).

Based on these observations, it is reasonable to speculate that the structural differences may result in distinct electrochemical behaviors. The spherical morphology of NMC811 promotes advanced packing density, whereas the flake-like layered structure of Li-rich-NMC-SiMg may provide higher surface area and shortened $\mathrm{Li}^+$ diffusion pathways, potentially leading to higher capacity.

We are currently conducting long-term cycle life testing of Li-rich-NMC-SiMg to further assess its commercial potential.

\newpage
\clearpage

\clearpage
\newpage
\vspace*{-1cm}

\clearpage
\newpage
\vspace*{-1cm}

\clearpage

\section{Synthesis of NMC-SiMg, NMC-SiCa, NMC-MgB, and Li-rich-NMC-SiMg}
\label{sec:wet_lab_synthesis}

\begin{figure}[ht]
    \centering
    \includegraphics[width=\linewidth]{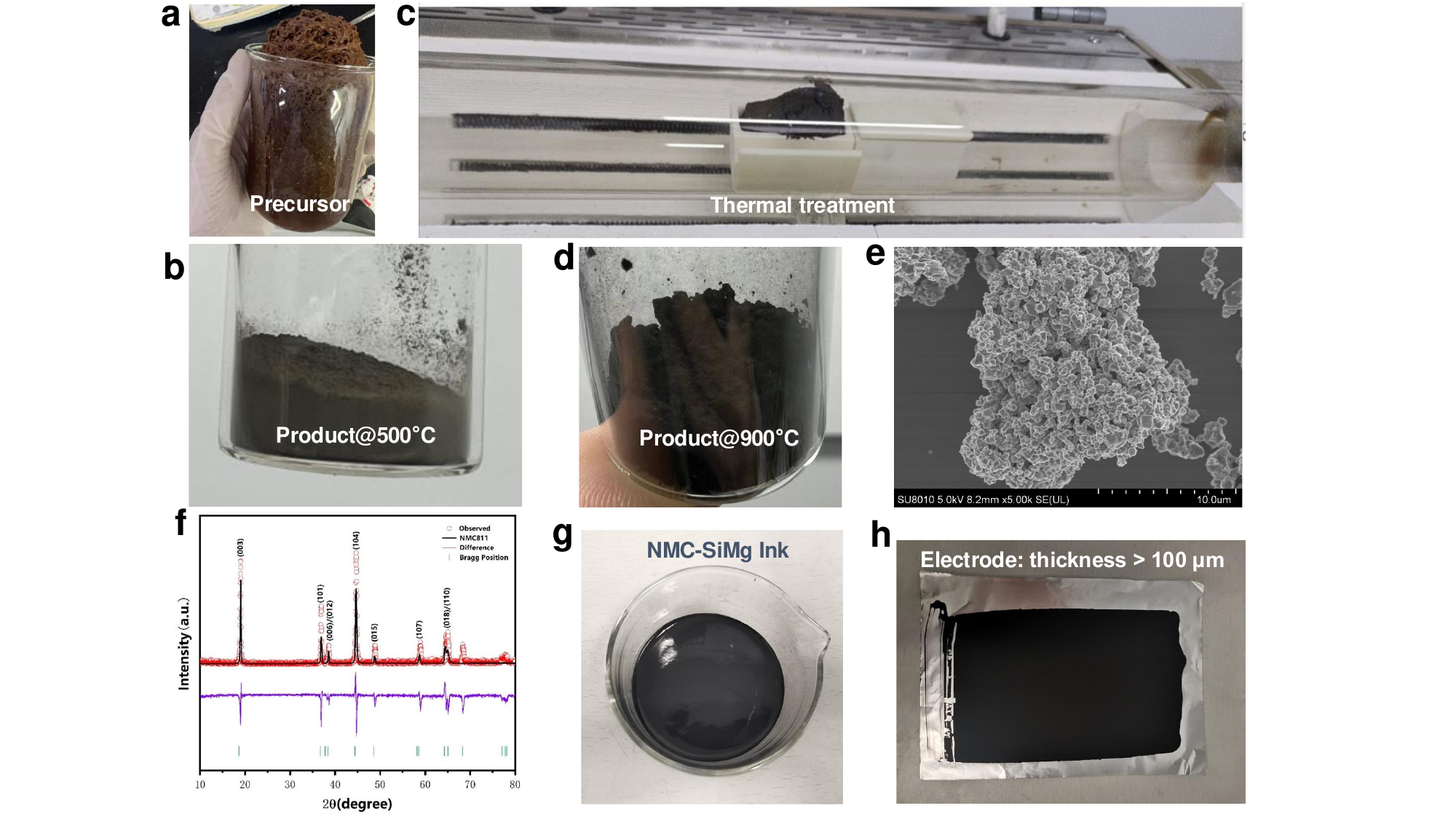} 
    \caption{\small \textbf{(a) The optical image of precusor of NMC-SiMg; (b) the product after heat-treatmemt @ 500 °C; (c) the thermal treatment process of NMC-SiMg production; (d) the final product of NMC-SiMg after heat-treatmemt @ 900 °C; (e) the typical SEM image of NMC-SiMg; (f) the XRD patterns of NMC811 and NMC-SiMg; (g) the NMC-SiMg ink for coating; (h) the NMC-SiMg electrode on Al foil with a thickness larger than 100 micro meters.}}
    \label{fig:SF6}
\end{figure}

All the NMC811-derived samples were synthesized based on output data following the procedure detailed in the Supplementary Materials section. The synthesis process of the representative, NMC-SiMg candidate, is illustrated in~\Cref{fig:SF6}, which presents a step-by-step progression from precursor to the final NMC-SiMg product through optical imaging. The NMC-SiMg precursor, obtained via a sol-gel approach, exhibits a grey color (\Cref{fig:SF6}a). Following heat treatment at 500 °C, it transitions to a deeper grey (\Cref{fig:SF6}b). The heat-treatment setup is depicted in ~\Cref{fig:SF6}c, while the final product, obtained after thermal processing at 900 °C, appears as a deep black powder (\Cref{fig:SF6}d). The SEM image (\Cref{fig:SF6}e) reveals nanoparticles at the nanoscale, and XRD patterns (\Cref{fig:SF6}f) indicate that while NMC-SiMg retains a similar crystal structure to NMC811, slight deviations are observed (\Cref{fig:SF6}g), which will be discussed in detail below. The NMC-SiMg ink (\Cref{fig:SF6}h), formulated by mixing PVDF and conductive carbon in NMP, exhibits optimal viscosity for coating. Consequently, the coated NMC-SiMg electrode, with a thickness exceeding 100 $\mu$m, demonstrated strong adhesion to aluminum foil, making it well-suited for industrial lithium-ion battery applications. 

The appropriate proportions of precursors, including metal ion acetates (Mn, Co, Ni), magnesium nitrate (Mg(NO\textsubscript{3})\textsubscript{2}), or calcium nitrate (Ca(NO\textsubscript{3})\textsubscript{2}) lithium hydroxide (LiOH), and silicic acid (H\textsubscript{2}SiO\textsubscript{3}), or boric acid (H\textsubscript{3}BO\textsubscript{3}), were dissolved in deionized water following the stoichiometric ratio, with magnesium and lithium salts in 5\% excess. Citric acid was used as a chelating agent, maintaining a molar ratio of metal ions to citric acid at 1:1.2. The mixture underwent ultrasonic mixing until fully dissolved. The reaction was then sealed and maintained at 55 °C for 12 hours before being evaporated to a brown-green gel at 80 °C in an oil bath. The resulting gel was treated at 180 °C for 10 hours under ventilation to form a solid, which was then ground into powder. For heat treatment, the powder was preheated in a tube furnace at 500 °C for 5 hours, followed by calcination at 900 °C for 20 hours in an air atmosphere to obtain the final product.

\subsection{Synthesis of NMC811}
The NMC811 was prepared using a similar standard protocol to that of NMC-SiMg, except without the addition of magnesium nitrate (Mg(NO\textsubscript{3})\textsubscript{2}), calcium nitrate (Ca(NO\textsubscript{3})\textsubscript{2}), boric acid (H\textsubscript{3}BO\textsubscript{3}), and silicic acid (H\textsubscript{2}SiO\textsubscript{3}). The final product, NMC811, which has a lower lithium stoichiometric ratio and lacks magnesium and silicon, was used for comparison.

\subsection{Electrode ink and Coin Cells making}
The NMC-SiMg ink was prepared using a standard protocol. First, polyvinylidene fluoride (PVDF, Sigma-Aldrich, 99\%) was dissolved in NMP at 60 °C. Then, NMC-SiMg, conductive carbon (Super P), and PVDF were dispersed in NMP at a mass ratio of 8:1:1 under continuous stirring with a magnetic stirrer (Camlab) at 60 °C for two hours. After thorough incorporation and mixing of the active NMC-SiMg material, the resulting slurry was coated onto aluminum foil (MSE Supplies) at a wet gap thickness of 30 µm and dried at 60 °C for approximately 30 minutes. Once dried, the electrodes were cut into 14 mm circular discs and assembled into coin cells within an inert glove box environment using standard coin cell casings (MTI, Al-Clad CR2032).

For all the coin cells, Celgard 2000 separators, lithium chips (MTI; thickness: 0.6 mm, diameter: 16 mm, purity: 99\%), a single wave spring (MTI, Al-Clad stainless steel for CR2032/CR2016 casings), two stainless steel spacers (MTI; 15.8 $\times$ 1.0 mm), and 60 µL of electrolyte were used. All the procedures were conducted in the glovebox filled with Ar, with O2 and H2O contend less than 0.1 ppm.

\subsection{Physicochemical Properties}
To assess the stability of NMC-SiMg in various common solvents, 1 g of the material was immersed in NMP solutions for 24 h, followed by complete solvent evaporation. The samples were then sealed with Kapton tape inside a glove box to prevent exposure to air. X-ray diffraction (XRD, Stoe STADI-P) was performed on the powder samples (sealed with Kapton tape) using Mo K$\alpha$ radiation ($\lambda$ = 0.0709 nm), with data collected over a 2$\theta$ range of 10–80°. 

\subsection{Electrochemical Testing}
Electrochemical characterizations were performed on coin cells to assess their performance under various conditions. All the electrochemical characterization was performed using a BioLogic battery test system (BCS-805, BioLogic). Cell cycling was conducted within a voltage range of 2.6–4.3 V, with an initial cycle at 0.1C (1C = 300 mA g$^{-1}$). Potentiostatic EIS measurements were performed using a Gamry potentiostat (Interface 1010E), with a frequency range of 0.1–1.0 × 10$^6$ Hz and a voltage perturbation of 5 mV.

\end{document}